%% file: acl_latex.tex
\pdfoutput=1
\PassOptionsToPackage{table}{xcolor}

\documentclass[11pt]{article}
\newif\iflongpaper

\longpapertrue 
\newcommand{\shortlong}[2]{%
  \iflongpaper
    #2%
  \else
    #1%
  \fi
}

\usepackage[preprint]{acl}

\usepackage{times}
\usepackage{latexsym}

\usepackage[T1]{fontenc}
\usepackage[utf8]{inputenc}

\usepackage{hyperref}
\usepackage{adjustbox}
\usepackage{graphicx}

\usepackage{microtype}
\usepackage{booktabs,arydshln}
\usepackage{array}
\usepackage{ragged2e}
 \usepackage{amsmath}
 \usepackage{threeparttable}
 \usepackage{multicol}
 \usepackage{multirow}
 \usepackage{ccicons}
 \usepackage{chngcntr}

 \makeatletter
\def\adl@drawiv#1#2#3{%
        \hskip.5\tabcolsep
        \xleaders#3{#2.5\@tempdimb #1{1}#2.5\@tempdimb}%
                #2\z@ plus1fil minus1fil\relax
        \hskip.5\tabcolsep}
\newcommand{\cdashlinelr}[1]{%
  \noalign{\vskip\aboverulesep
           \global\let\@dashdrawstore\adl@draw
           \global\let\adl@draw\adl@drawiv}
  \cdashline{#1}
  \noalign{\global\let\adl@draw\@dashdrawstore
           \vskip\belowrulesep}}
\makeatother

\makeatletter
\def\adl@drawiv#1#2#3{%
  \hskip.5\tabcolsep
  {\color[gray]{0.7} 
  \xleaders#3{#2.5\@tempdimb #1{1}#2.5\@tempdimb}%
      #2\z@ plus1fil minus1fil\relax}%
  \hskip.5\tabcolsep}
\newcommand{\cdashlinelrg}[1]{%
  \noalign{\vskip\aboverulesep
           \global\let\@dashdrawstore\adl@draw
           \global\let\adl@draw\adl@drawiv}
  \cdashline{#1}
  \noalign{\global\let\adl@draw\@dashdrawstore
           \vskip\belowrulesep}}
\makeatother

\usepackage{pifont}
\usepackage{fontawesome5}

\definecolor{richgreen}{RGB}{102, 204, 102}
\definecolor{richred}{RGB}{255, 99, 71}

\newcommand{\cmark}{\textcolor{richgreen}{\ding{51}}}
\newcommand{\xmark}{\textcolor{richred}{\ding{55}}}
\newcommand{\smark}{\textcolor{blue}{$\approx$}}
\newcommand{\dmark}{\textcolor{blue}{$\downarrow$}}

\definecolor{richorange}{RGB}{255, 165, 0}

\definecolor{ctxorange}{RGB}{217,95,2} 
\definecolor{ctxteal}{RGB}{27,158,119} 
\definecolor{coolgray}{RGB}{160,160,170}

\newcommand{\rmark}{\textcolor{richorange}{\faExclamationTriangle}}

\definecolor{slateblue}{RGB}{100,140,200}

\newcommand{\incmark}{\textcolor{slateblue}{\faAdjust}}

\newcommand{\qmark}{\textcolor{gray!80}{\faQuestion}}

\newcommand{\zeroshotmark}{\textcolor{orange}{\faBolt}}

\newcommand{\firemark}{\textcolor{richred}{\faFire}}
\newcommand{\snowmark}{\textcolor{slateblue}{\faSnowflake}}

\definecolor{augpurple}{RGB}{117,107,177}
\newcommand{\augmark}{\textcolor{augpurple}{\faClone}}

\usepackage[most]{tcolorbox}   
\usepackage{siunitx}           
\newcommand{\pill}[2]{%
  \tcbox[
    colback=#1!12,
    colframe=#1!55!black,
    boxrule=0.25pt,
    arc=2pt,
    left=3pt,right=3pt,top=1pt,bottom=1pt,
    boxsep=0pt,
    valign=center,
    before skip=0pt,after skip=0pt,
    on line
  ]{\scriptsize #2}%
}

\providecommand{\truncctx}{\textcolor{ctxorange}{\faCompress}}
\providecommand{\fullctx}{\textcolor{ctxteal}{\faInfinity}}

\newcommand{\secbadge}[1]{\pill{coolgray}{\faClock\ \SI{#1}{\second}}}
\newcommand{\fullbadge}{\pill{ctxteal}{Full}}

\providecommand{\inputtrunc}[1]{\truncctx\ \secbadge{#1}}
\providecommand{\inputfull}{\fullctx\ \fullbadge}

\usepackage{colortbl} 

\newcolumntype{R}{p{0.05cm}} 

\newcolumntype{P}[1]{>{\RaggedRight\arraybackslash}p{#1}}
\usepackage{makecell}
\newcolumntype{C}[1]{>{\centering\arraybackslash}p{#1}}
\newcolumntype{D}[1]{>{\raggedright\arraybackslash}p{#1}}

\usepackage{amsmath}

\usepackage{inconsolata}

\usepackage{graphicx}

\usepackage{afterpage}

\usepackage[capitalize]{cleveref}
\crefname{figure}{Figure}{Figures}
\crefname{table}{Table}{Tables}

\title{\textit{Summarizing Speech:} A Comprehensive Survey}

\newcommand{\authorsep}{\quad}

\author{%
\textbf{Fabian Retkowski}$^1$\authorsep
\textbf{Maike Züfle}$^1$\authorsep
\textbf{Andreas Sudmann}$^2$\authorsep
\textbf{Dinah Pfau}$^3$\\
\textbf{Shinji Watanabe}$^4$\authorsep
\textbf{Jan Niehues}$^1$\authorsep
\textbf{Alexander Waibel}$^{1,4}$\\
\textsuperscript{1}KIT\authorsep
\textsuperscript{2}University Bonn\authorsep
\textsuperscript{3}Deutsches Museum\authorsep
\textsuperscript{4}CMU\\
\texttt{\{fabian.retkowski,maike.zuefle,jan.niehues,alex.waibel\}@kit.edu} \\
\texttt{asudmann@uni-bonn.de}\authorsep 
\texttt{d.pfau@deutsches-museum.de}\authorsep
\texttt{swatanab@andrew.cmu.edu}
}

\begin{document}
\maketitle

\input{00_abstract}
\input{01_introduction_motivation}

\input{02_challenges}

\input{03_problem_form}
\input{04_data}
\input{05_eval}
\input{06_approaches}
\input{07_challenges}

\input{08_conclusion_limitation}

\shortlong{
\section*{Acknowledgments}
This research is supported by the project "How is AI Changing Science? Research in the Era of Learning Algorithms" (HiAICS), funded by the Volkswagen Foundation, and partially by the European Union’s Horizon research and innovation programme under grant agreement No. 101135798, project Meetween (My Personal AI Mediator for Virtual MEETtings BetWEEN People).
}{\section*{Acknowledgments}
This research is supported by the project "How is AI Changing Science? Research in the Era of Learning Algorithms" (HiAICS), funded by the Volkswagen Foundation, and partially by the European Union’s Horizon research and innovation programme under grant agreement No. 101135798, project Meetween (My Personal AI Mediator for Virtual MEETtings BetWEEN People).}

\bibliography{custom, anthology_0, anthology_1, custom_manual}

\newpage

\appendix
\counterwithin{table}{section}
\renewcommand{\thetable}{\thesection\arabic{table}}
\counterwithin{figure}{section}
\renewcommand{\thefigure}{\thesection\arabic{figure}}
\newpage
\clearpage
\shortlong{\input{appendices/history_appendix}}{}
\input{appendices/datasets_appendix}
\input{appendices/eval_appendix}

\input{appendices/models_appendix}
\input{appendices/venues_appendix}

\end{document}

%% file: 00_abstract.tex
\begin{abstract}
Speech summarization has become an essential tool for efficiently managing and accessing the growing volume of spoken and audiovisual content. However, despite its increasing importance, speech summarization remains loosely defined. The field intersects with several research areas, including speech recognition, text summarization, and specific applications like meeting summarization. This survey not only examines existing datasets and evaluation protocols, which are crucial for assessing the quality of summarization approaches, but also synthesizes recent developments in the field, highlighting the shift from traditional systems to advanced models like fine-tuned cascaded architectures and end-to-end solutions. In doing so, we surface the ongoing challenges, such as the need for realistic evaluation benchmarks, multilingual datasets, and long-context handling.
\end{abstract}

%% file: 01_introduction_motivation.tex
\section{Introduction}

The digital age is increasingly shaped by the high volume of spoken and audiovisual content, diverging from text-centric origins. Podcasts now number in the millions, with over 500 million global listeners and up to 30 million new episodes released per year \cite{litterer_mapping_2024,listennotes2025}. Platforms like YouTube and TikTok receive hundreds of thousands of hours of video every minute, a flood of content growing exponentially since the early 2000s and far outpacing human attention and capacity \cite{ceci_hours_2024}. Meanwhile, everyday communication is shifting from text to voice, with users sending over 7 billion voice messages daily via apps like WhatsApp \cite{whatsapp_making_2022}. 

\begin{figure}[tp]
    \centering
    \includegraphics[width=\linewidth]{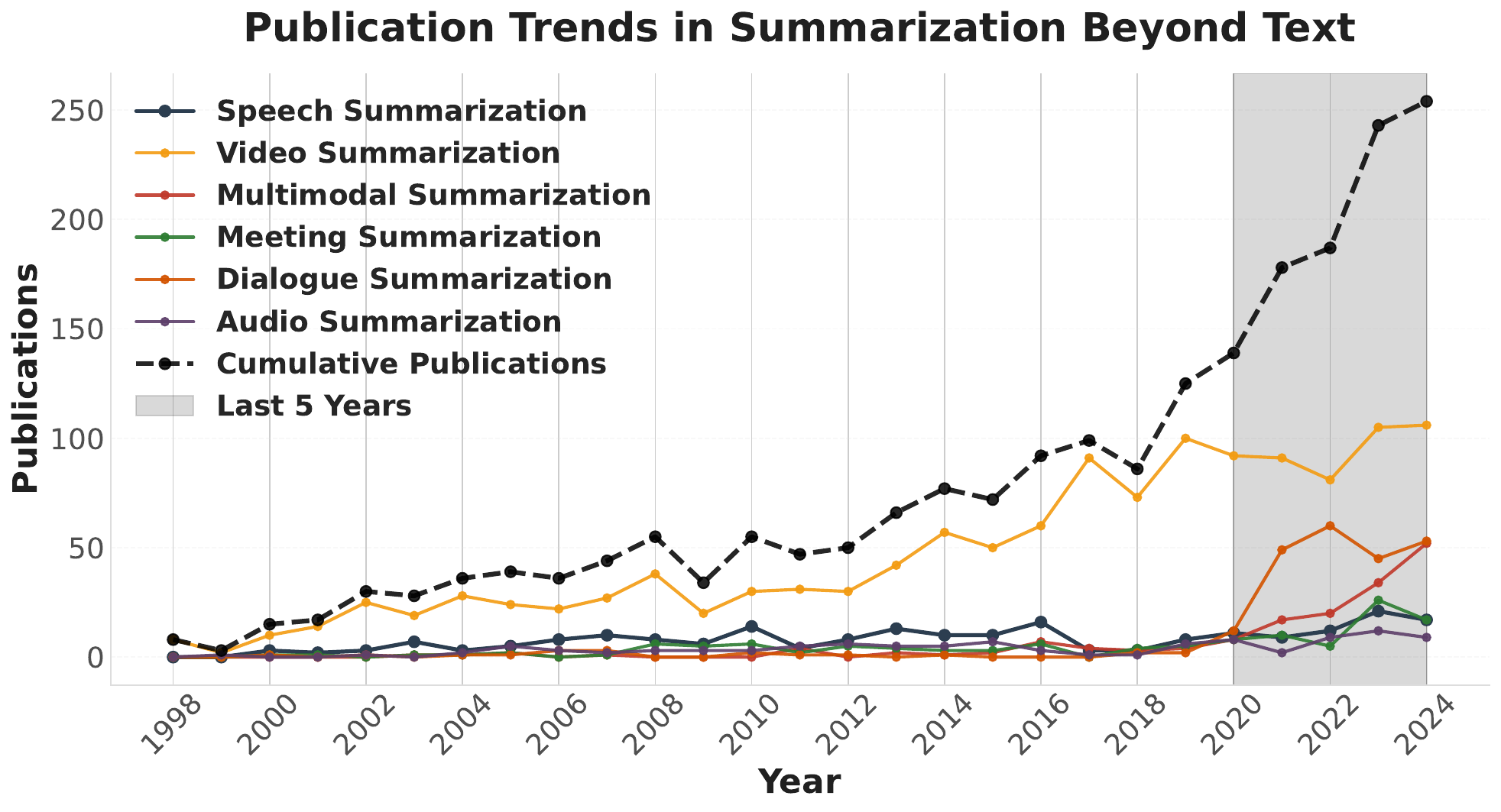}
    \caption{Publication trends in summarization beyond text, based on search results from dblp.org, showing significant growth and evolving research focus.}
    \label{fig:publication-trends}\shortlong{\vspace{-0.35cm}}{}
\end{figure}

 But as audiovisual content becomes central to both media consumption and daily communication in the digital era, the resulting overload of speech data creates challenges for access, navigation, and comprehension \cite{ghosal_report_2022}. In response, \textit{speech summarization} (SSum) has emerged as a crucial way to make spoken content more manageable, enabling quicker information access, aiding research, and supporting everyday use across personal and professional contexts \cite{murray_generating_2010,li_hierarchical_2021,jung_interactive_2023}. Yet despite its growing relevance, SSum remains surprisingly underdefined, occupying a unique interdisciplinary position that has not yet been fully explored \cite{rezazadegan_automatic_2020,ghosal_report_2022}. \cref{fig:publication-trends} reveals an interesting tension in the field: while publication counts are modest compared to video summarization, SSum exists at the intersection of multiple thriving research areas, including \textit{automatic speech recognition} (ASR), \textit{text summarization} (TSum), and domain-specific applications like \textit{meeting summarization}. This is also evident in the publication distribution across different venues (see \cref{fig:venues_over_years}). This ambiguity in definition is both a challenge and an opportunity. SSum is not merely the application of TSum to ASR output, nor simply the audio component of video summarization. It requires addressing distinctive complexities, including disfluencies, prosody, speaker dynamics, and contextual elements \cite{zhu_hierarchical_2020,song_towards_2022,sharma-etal-2024-speech}. The field's fragmentation across research communities has led to parallel developments that would benefit from unification. From meeting summarization \cite{rennard_abstractive_2023} to podcast summarization \cite{jones_trec_2020} to multimodal summarization \cite{jangra_survey_2023}, all tackle speech content but often operate in isolation, using different methodologies and benchmarks. This creates a critical need for survey work that brings these interconnected domains together and identifies broader challenges.

\shortlong{}{
\subsection{Early Work}
\input{sections/history}
}

\subsection{Scope of the Survey}

\shortlong{}{\paragraph{Survey Focus and Scope.} }This survey provides a synthesis of the evolving landscape of SSum, bridging fragmented developments across ASR, TSum, dialogue summarization, and multimodal applications. Our primary focus is on work published since 2020, reflecting rapid transformation of the field since then. The most recent survey prior to this work by \citet{rezazadegan_automatic_2020} captured pre-2020 approaches, largely \shortlong{}{based on} traditional pipelines and early neural models. In the years following, the field has shifted: cascaded systems now leverage fine-tuned encoder-decoder (ED) models, prompting or adapting LLMs has become common, and end-to-end (E2E) models are increasingly explored. \shortlong{For historical context, a concise overview of earlier work is provided in \cref{app:history}.}{} Unlike prior surveys on meeting~\cite{rennard_abstractive_2023}, dialogue~\cite{tuggener_are_2021,kirstein_cads_2025}, text~\cite{gambhir_recent_2017,el-kassas_automatic_2021,sudmann_current_2023}, and multimodal summarization~\cite{jangra_survey_2023}, this work focuses specifically on \textit{spoken language} as input and \textit{text} as output (i.e., \textit{speech-to-text summarization}) across diverse application domains while clearly delineating the scope of SSum from neighboring fields like video summarization.

\shortlong{}{\paragraph{Structure and Chapter Overview.} The survey is organized thematically, first outlining the challenges unique to speech processing in general (\cref{sec:challenges}), then formalizing the problem in the broader landscape and its input and output modalities (\cref{sec:problems}), followed by a detailed examination of available data resources (\cref{sec:data}), evaluation methods (\cref{sec:eval}), core modeling approaches (cascaded, LLM-based, and end-to-end in \cref{sec:approaches}), and concluding with future directions (\cref{sec:future}).}


%% file: sections/history.tex
In the 20th century, advances in telecommunications, military research, and information technology laid the foundations for speech processing. While early summarization efforts focused on textual data \cite{luhn_automatic_1958}, the challenge of summarizing speech gained prominence later. ASR began to mature in the 1980s and 1990s, particularly through statistical methods based on Markov models \cite{baum_maximization_1970, jelinek_continuous_1976, rabiner_tutorial_1989} and connectionist models \cite{waibel_phoneme_1989, franzini_connectionist_1990, renals_connectionist_1994}, laying the groundwork for processing speech.
In the 1990s, data-driven methods increasingly linked ASR and natural language processing (NLP), with early projects highlighting the potential of summarization for large-scale spoken content and identifying challenges specific to spontaneous speech, such as topic drift, disfluencies, hesitations, and ASR errors through corpora like Switchboard \cite{godfrey_switchboard_1992} and programs like TIPSTER \cite{suhm94_icslp, zeppenfeld1997recognition, gee_tipster_1998}. Around 2000, research on SSum gained traction, initially adapting TSum via extractive methods for challenges like telephone dialogues \cite{zechner_diasumm_2000, mckeown_text_2005} and broadcast news \cite{hori2002automatic}, selecting salient segments. Concurrently, early multimodal approaches were explored for complex meeting interactions \cite{yang1998visual,gross2000towards} culminating in the development of rich, annotated corpora such as AMI \cite{carletta_ami_2006} and ICSI \cite{janin_icsi_2003}, foundational for meeting summarization. By the mid-2000s, extractive systems increasingly relied on features specific to speech, including prosody, speaker activity, and dialog acts \cite{koumpis_automatic_2005, maskey_comparing_2005, murray_extractive_2005}. Early work raised questions about how to evaluate summaries of spoken language in the presence of ASR errors and disfluencies \cite{whittaker_scan_1999, zechner_minimizing_2000}. In subsequent years, evaluation became standardized through ROUGE \cite{lin-2004-rouge}. Finally, early steps toward abstractive SSum also emerged through a combination of speech paraphrasing and sentence compression techniques \cite{hori_speech_2003}.  Over the following two decades, extractive methods remained dominant, but the adoption of abstractive techniques steadily grew \cite{rezazadegan_automatic_2020}, driven by deep learning advances that enabled more fluent generation. Today, after encoder-decoder architectures and pretrained language models emerged, abstractive methods have become dominant in SSum \cite{rennard_abstractive_2023}. This shift also reflects user preferences, as humans tend to favor abstractive summaries for speech content \cite{murray_generating_2010}.

%% file: 02_challenges.tex
\section{Challenges of Speech Processing}
\label{sec:challenges}


\paragraph{Orality and Linguistic Variability.}  
Unlike written text, spoken language lacks structural markers such as punctuation, headings, or paragraph breaks \citep{rehbein-etal-2020-improving}, making it harder to detect topical shifts and organize content \citep{zechner_diasumm_2000, khalifa-etal-2021-bag}. Furthermore, speech often includes disfluencies and false starts \citep{khalifa-etal-2021-bag,kirstein-etal-2024-whats,teleki_etal_2024_quantifying}\shortlong{}{, complex speaker interactions and coreference links \cite{liu-etal-2021-coreference},} and features accents, dialects, and code-switching \citep{keswani2021dialects}, all of which add complexity. Prosodic features like intonation, rhythm, and emphasis also carry meaning \citep{aldeneh-etal-2021-learning} but are often lost in ASR-based pipelines. Finally, speech is often lengthy, unstructured, and semantically sparse, with important information scattered across speaker turns and interleaved with filler or redundant speech, making long-context modeling critical \cite{liu_topic-aware_2019}.

\paragraph{Acoustic Environment.}  
External acoustic factors such as overlapping speakers or background noise (e.g., applause or sound effects) are common in spoken content. These factors can either contribute valuable context or introduce noise \citep{jiminez2020}, posing challenges for systems that risk discarding useful cues or being disrupted by extraneous sounds \citep{cornell2023chime}.

\paragraph{Modality Constraints.}  
SSum presents notable technical challenges. First, real-world speech (e.g., meetings, lectures) often spans long durations, which can strain memory and processing resources  \citep{kumar2022meetingsummarizationsurveystate}. Second, many pipelines rely on ASR, and transcription errors introduce noise into downstream processing \citep{rennard_abstractive_2023, chowdhurytranscripterrors2024}. 

%% file: 03_problem_form.tex
\section{Problem Formulation}
\label{sec:problems}
\subsection{Speech Summarization}

\textit{Speech summarization} is the process of condensing spoken content into a shorter version while preserving essential information. It is most commonly understood as a \textit{cross-modal} task, where an audio signal (speech) is transformed into a textual summary (\textit{speech-to-text summarization, STT}). However, it is often implemented as a \textit{cascaded} approach, where an ASR system first transcribes the speech into text, followed by \textit{unimodal} \textit{text summarization} systems. Alternatively, the input may be a manually created transcript, in which case the summarization remains a form of speech summarization but is entirely text-based. The output can be either \textit{extractive}, where key sentences or phrases are directly taken from the original speech, or \textit{abstractive}, where the summary is generated in a rephrased form - the dominant approach in contemporary systems. It is notable that summarization can be performed at different granularities, such as sentence-level, segment-level, or document-level.

\subsection{Input Data Modalities}

\shortlong{}{\begin{figure*}[!htb]
    \centering 
    \includegraphics[trim=8mm 15mm 8mm 20mm, clip, width=0.75\textwidth]{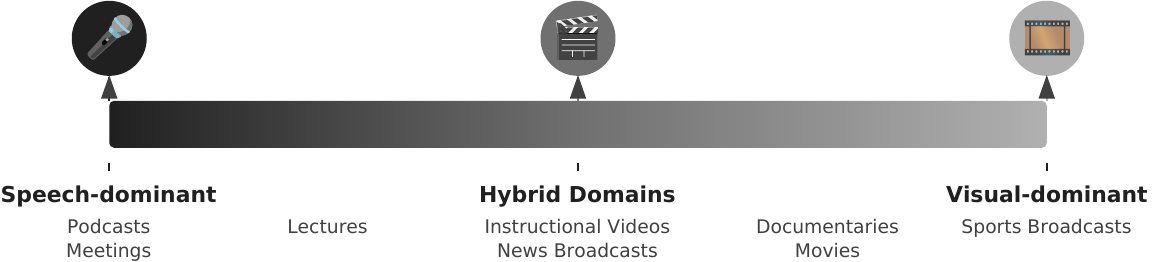}
    \caption{The Speech-Video Modality Importance Spectrum}
    \label{fig:multimodal_spectrum}
    
\end{figure*}
}

The input can take the form of raw audio or transcripts, either generated via ASR or created manually by humans. Similar trends have been observed in both human and automated summarization: the choice of input modality significantly impacts summary quality. For instance, \citet{sharma-etal-2024-speech} analyzes human-written summaries and finds that presenting annotators with raw speech, rather than transcripts, leads to more selective and factually consistent outputs. They also show that ASR errors reduce the informativeness and coherence of summaries. In parallel, incorporating speech-specific features such as prosody \shortlong{}{or speaker information} into SSum systems has been shown to improve performance \cite{inoue_improvement_2004,liu_hierarhical_2018}. For cascaded systems, the quality of ASR transcripts remains a limiting factor, with clear performance gaps compared to manual transcripts \cite{kano_attention-based_2021,binici_medsage_2025}.


\subsection{Applications and Related Tasks}

\subsubsection{Core Applications}

A core application of \shortlong{SSum}{speech summarization} is \textit{meeting summarization}, condensing free-form discussions into concise overviews, which can range from high-level summaries \cite{janin_icsi_2003,carletta_ami_2006} to more structured outputs like meeting minutes \cite{nedoluzhko_elitr_2022,hu_meetingbank_2023} or action item lists \cite{purver_detecting_2007,mullenbach_clip_2021,asthana_summaries_2024}, blurring the lines between summarization and structured information logging \cite{tuggener_are_2021}. More broadly, this falls under the umbrella of \textit{dialogue summarization}, which includes not only spoken interactions such as meetings, customer service calls, and interviews but also text-based dialogues like chat transcripts. Other prominent \shortlong{applications}{application domains} include \textit{podcast summarization} \cite{clifton_100000_2020,song_towards_2022} and \textit{presentation summarization}, which focuses on structured, monologic content such as lectures \cite{miller_leveraging_2019,lv_vt-ssum_2021,xie_using_2025}, TED Talks \cite{kano_attention-based_2021,shon_slue_2023}, and conference presentations \cite{zufle_nutshell_2025}. A further core area is \textit{YouTube video summarization}, which has emerged as a major testbed for SSum systems \cite{sanabria_how2_2018,retkowski_text_2024,qiu_mmsum_2024}. It encompasses a wide variety of content types, ranging from educational videos to interviews, vlogs, and news broadcasts, and poses unique challenges due to its diversity.


\subsubsection{Related Tasks}

\paragraph{Smart Chaptering.} Many speech summarization applications benefit from \textit{smart chaptering} (or topic segmentation), where spoken content is divided into coherent sections. This approach enables more granular summarization at the chapter level, while the chapter titles function as extreme summaries \cite{zechner_diasumm_2000,banerjee_generating_2015,ghazimatin_podtile_2024,retkowski_text_2024,xie_using_2025}. 

\paragraph{Subtitle Compression.} At an even finer granularity, \textit{sentence-wise SSum} \cite{matsuura_sentence-wise_2024} focuses on condensing individual spoken sentences into more concise forms. This task is particularly relevant to \textit{subtitle compression}, where subtitles may initially be transcriptions or translations of speech that are too long to fit on screen\hspace{0.1cm} or to be read comfortably by viewers. 
The task of subtitle compression 
addresses this by automatically shortening subtitle text while preserving its meaning \cite{liu_adapting_2020,papi-etal-2023-direct-speech,jorgensen_cross-lingual_2025,retkowski_zero-shot_2025}.

\shortlong{\paragraph{Adjacent STT Tasks.} Finally, SSum sits on a continuum with adjacent STT tasks such as spoken QA, ASR, and speech translation, see \Cref{app:adjacent}.}{\paragraph{Adjacent Speech-to-Text Tasks.} \input{sections/adjacent_stt}} 


\subsubsection{Additional Input Modalities}

\shortlong{\begin{figure}[!t]
    \centering 
    \includegraphics[trim=4mm 15mm 6mm 15mm, clip, width=0.48\textwidth]{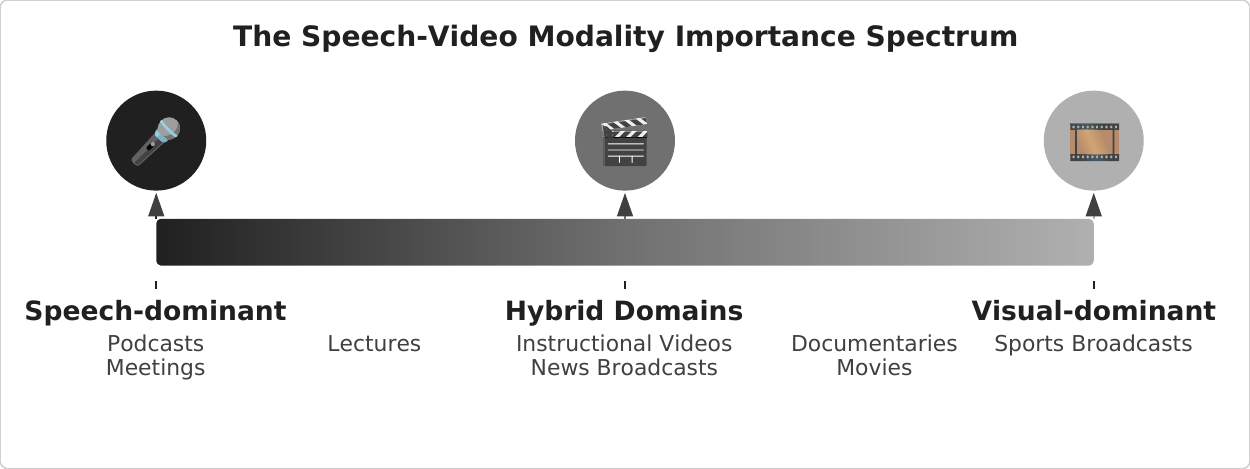}
    \caption{The Speech-Video Modality Spectrum}
    \label{fig:multimodal_spectrum}\shortlong{\vspace{-0.3cm}}{}
\end{figure}}{}

\paragraph{The Value of Visual Cues.} Speech summarization inherently extends into \textit{multimodal summarization} as speech is frequently embedded within environments rich with complementary visual and contextual information. Multimodal information has been shown to provide significant value to many SSum systems. For example, incorporating modalities beyond text or audio has been demonstrated to enhance summarization of instructional videos \cite{palaskar_multimodal_2019,khullar_mast_2020} while non-verbal cues like eye gaze, speaker focus, and head orientation improve meeting summarization \cite{nihei_fusing_2018,li_keep_2019}. Reflecting this, many datasets used in SSum, such as How2 \cite{sanabria_how2_2018} or AMI \cite{carletta_ami_2006}, provide not only audio but also video.

\paragraph{The Continuum Between Speech and Video Summarization.} This connection highlights a spectrum between SSum and \textit{video summarization} (visualized in \cref{fig:multimodal_spectrum}). While speech-focused approaches treat visuals as complementary, true video summarization considers visual elements essential rather than supplementary. Different domains fall along this continuum: podcasts and meetings represent speech-dominant contexts where non-verbal cues primarily contextualize speech, while sports broadcasts and action-rich movies sit at the visual-dominant end where visual composition and action sequences carry critical narrative information. 

\subsubsection{Beyond Text as Output Modality}
While this survey primarily addresses speech-to-text summarization, we also want to discuss alternative or additional output modalities briefly. Early work by \citet{furui_ssum_2004} introduced a cascaded \textit{speech-to-speech summarization} approach, where speech was first transcribed, summarized textually, and then synthesized back into audio. More recently, ESSumm \cite{wang_essumm_2022} has bypassed transcripts entirely, selecting salient audio segments directly. \shortlong{}{Closely related tasks include \textit{podcast preview extraction} \cite{zhu_transforming_2025}, where systems select engaging contiguous segments of speech to serve as previews.} Visual outputs have also been explored under tasks like \textit{multimodal summarization with multimodal output} (MSMO), where systems generate both textual summaries and representative visual thumbnails \cite{zhu_msmo_2018,qiu_mmsum_2024}.



%% file: sections/adjacent_stt.tex
Highly abstractive STT tasks like \textit{spoken question answering} \cite{chuang20b_interspeech} and \textit{qualitative coding} of speech \cite{retkowski-etal-2025-ai} exhibit SSum-like processes, abstracting and distilling core information. More broadly, many STT tasks share conceptual overlap with SSum, differing in their level of abstraction. For example, ASR frequently incorporates disfluency removal and sentence restructuring to improve readability \cite{jamshid_lou_johnson_2020_end}, while \textit{speech translation} rephrases spontaneous speech across languages, often requiring significant abstraction to handle idiomatic expressions and cultural references \cite{gaido_etal_2024_speech}.

%% file: 04_data.tex
\section{Data Resources}
\label{sec:data}

\cref{tab:datasets} presents datasets relevant to speech summarization and related tasks\footnote{An up-to-date interactive version of this dataset table is available at \url{https://ssum-survey.github.io/datasets}.}. Given the scarcity of dedicated SSum datasets with true summaries, we also include datasets that rely on surrogate summaries (discussed below) as well as text-to-text summarization datasets if they are based on spoken content or \shortlong{}{closely} resemble speech in structure and style. \shortlong{}{Subtitle compression serves as a fine-grained form of summarization, while segmentation can involve either segment-level summaries or extreme summarization, such as generating short titles.}

\input{tables/dataset_table}

\paragraph{Limitations of Surrogate Summaries.} Many SSum datasets rely on 
\textit{surrogate summaries}, such as creator descriptions (e.g., from YouTube videos and podcast episodes; \citealt{sanabria_how2_2018, clifton_100000_2020}), or paper abstracts \cite{liu_what_2025,zufle_nutshell_2025}. While these summaries provide a convenient source of training data, they were not originally designed as true summaries, leading to several limitations. First, surrogate summaries are often of poor quality because they \shortlong{}{typically} serve a different purpose: descriptions \shortlong{act}{function} as teasers, abstracts follow distinct stylistic conventions. \citet{manakul_podcast_2022} highlight this issue by evaluating \shortlong{}{the quality of} creator-provided descriptions in the \hyperlink{spotify-row}{Spotify Podcast Dataset}, finding that 26.3\% were rated as ``Bad''\shortlong{}{, while only 15.6\% were considered ``Excellent''}. Tellingly, automatic systems outperformed the original descriptions in quality \cite{manakul_cued_speech_2020}. Second, surrogate summaries may contain information not present in the original speech. \citet{zufle_nutshell_2025} found that while 70.0\% of paper abstracts were considered good summaries, 63.3\% included content absent from the talk. Likewise, in the \hyperlink{summscreen-row}{SummScreen} dataset, TV recaps incorporate visual context (actions, settings) missing from the transcript, leading to \shortlong{}{potential} content mismatches and \shortlong{}{model} hallucinations \cite{chen_summscreen_2022}.


\paragraph{Scarcity of Datasets.} Our overview illustrates that the field is characterized by inconsistent benchmarks, a lack of high-quality, large-scale datasets, and a landscape of fragmented, interrelated tasks and problems rarely contextualized in the broader field. This issue is further exacerbated by the fact that two of the most popular and largest datasets, namely \hyperlink{how2-row}{How2} and the \hyperlink{spotify-row}{Spotify Podcast Dataset}, are no longer publicly available to researchers.

\paragraph{Synthetic Data.} A promising approach to mitigate data volume limitations is synthesizing data, as shown in recent research. For example, in the context of speech summarization, several works \cite{matsuura_leveraging_2023,matsuura_sentence-wise_2024,eom_squba_2025} use a TTS system to generate synthetic speech input from text, while LLMs can be leveraged to generate reference summaries \citep{chen_floras,jung_augsumm_2024,le-duc_real-time_2024,eom_squba_2025}. Taking this further, LLMs have been leveraged to produce entire multi-party social conversations that achieve quality close to human-generated data \cite{chen_places_2023,suresh_diasynth_2025}. Additionally, LLMs have been employed to synthesize ASR errors, improving the robustness of summarization models \citep{binici_medsage_2025}, while traditional audio data augmentation, such as adding background noise or reverberation, remains valuable for E2E SSum \citep{home_assist_heab2019}.

\paragraph{Out-of-Domain Data.} Another strategy to overcome limited in-domain data is cross-domain pretraining, where models are first trained on large-scale text-based summarization datasets such as CNN/DailyMail, XSum, or \hyperlink{sam-row}{SAMSum}. These corpora help models acquire general summarization abilities before being fine-tuned on speech-specific datasets. This approach has been shown to improve performance on diverse speech summarization benchmarks, including long meeting summarization \citep{zhu_hierarchical_2020,zhang_exploratory_2021}.

\paragraph{Recommended Resources.} Given the limitations of current benchmarks, including the unavailability of widely used datasets and the small scale of others such as \hyperlink{ami-row}{AMI}, there is a clear need for viable alternatives. Among the available datasets, several stand out for their combination of \textit{accessible audio} and \textit{considerable scale}. \hyperlink{slue-row}{SLUE-TED}, \hyperlink{nutshell-row}{NUTSHELL} and \hyperlink{vista-row}{VISTA} offer high-quality speech aligned with abstractive summaries, based on TED talks and AI conference presentations. \hyperlink{ytseg-row}{YTSeg}, while using chapter titles as summaries, provides large-scale, manually transcribed YouTube content and is particularly well suited for long-context and structure-aware SSum. \hyperlink{meetingbank-row}{MeetingBank} complements these with long-form meetings and segment-level summaries. Several other datasets in \cref{tab:datasets} are also promising, especially when paired with synthetic speech via TTS to compensate for the lack of audio.



%% file: tables/dataset_table.tex
\begin{table*}[!ht]
\centering
\renewcommand{\arraystretch}{0.8}
\setlength{\aboverulesep}{1.5pt}
\setlength{\belowrulesep}{1.5pt}
\setlength{\tabcolsep}{0.4em}
\renewcommand{\baselinestretch}{0.95}
\newcommand{\customtoprule}{\specialrule{1pt}{0pt}{2pt}}
\newcommand{\custombottomrule}{\specialrule{1pt}{2pt}{0pt}}
\newcommand{\custommidrule}{\specialrule{0.3pt}{1.5pt}{1.5pt}}
\newcommand{\seprule}{\specialrule{1pt}{2pt}{2pt}}
\tiny
\begin{threeparttable}
\begin{tabular}{P{0pt} P{1.35cm} P{1.85cm} P{1.85cm} P{0.7cm} P{1.55cm} P{2.40cm} C{0.875cm} C{0.5cm} C{0.5cm} C{1.05cm}}
\customtoprule
& \textbf{Dataset} & \textbf{Reference} & \textbf{Domain} & \textbf{Lang.} & \textbf{Size} & \textbf{Summary Type} & \textbf{Transcript} & \textbf{Audio} & \textbf{Video} & \textbf{License}\\
\custommidrule
\rmark\tnote{c} 
& \makebox[0pt][l]{\hypertarget{how2-row}{}}\href{https://github.com/srvk/how2-dataset}{How2~\faExternalLink*} 
& \citet{sanabria_how2_2018} 
& Instructional videos (YouTube) 
& EN, PT\tnote{h} 
& 80k videos (2k hours) 
& Abstractive (video descriptions) 
& Manual 
& \dmark\tnote{a} 
& \dmark\tnote{a}  
& CC-BY-SA-4.0 \\
\custommidrule
& \makebox[0pt][l]{\hypertarget{ytseg-row}{}}\href{https://huggingface.co/datasets/retkowski/ytseg}{YTSeg~\faExternalLink*} 
& \citet{retkowski_text_2024} 
& YouTube videos (various types/topics) 
& EN 
& 19.3k videos (6.5k hours) 
& Abstractive (segment-based, chapter titles) 
& Manual 
& \cmark 
& \dmark\tnote{a} 
& CC-BY-NC-SA-4.0 \\
\custommidrule
& \href{https://github.com/Jielin-Qiu/MMSum_model}{MMSum~\faExternalLink*} 
& \citet{qiu_mmsum_2024} 
& YouTube videos (various types/topics) 
& EN 
& 5.1k videos (1.2k hours) 
& Abstractive (segment-based, chapter titles, thumbnails) 
& Manual 
& \dmark\tnote{a} 
& \dmark\tnote{a} 
& CC-BY-NC-SA \\
\custommidrule
& \makebox[0pt][l]{\hypertarget{floras-row}{}}\href{https://huggingface.co/datasets/espnet/floras}{FLORAS~50~\faExternalLink*} 
& \citet{chen_floras} 
& YouTube videos (various types/topics) 
& 50 
& 9.3k hours 
& Abstractive (synthetic LLM summaries) 
& Manual 
& \cmark
& \xmark
& CC-BY-3.0 \\
\custommidrule
& \href{https://github.com/Dod-o/VT-SSum}{VT-SSum~\faExternalLink*} 
& \citet{lv_vt-ssum_2021} 
& Lecture videos (VideoLectures.net) 
& EN 
& 9.6k videos 
& Abstractive (segment-based, slide text) 
& ASR 
& \dmark\tnote{a} 
& \dmark\tnote{a} 
& CC-BY-NC-ND-4.0 \\
\custommidrule
& \makebox[0pt][l]{\hypertarget{nutshell-row}{}}\href{https://huggingface.co/datasets/maikezu/nutshell}{NUTSHELL~\faExternalLink*} 
& \citet{zufle_nutshell_2025} 
& Conference talks (*ACL talks) 
& EN 
& 6.3k talks (1.2k hours) 
& Abstractive (paper abstracts) 
& \xmark
& \cmark 
& \dmark\tnote{a} 
& CC-BY-4.0 \\  
\custommidrule
\incmark\tnote{g}& \href{https://huggingface.co/FBK-MT/MCIF}{MCIF~\faExternalLink*} 
& \citet{papi2025mcifmultimodalcrosslingualinstructionfollowing} 
& Conference talks (*ACL talks) 
& EN, DE\tnote{i}, IT\tnote{i}, ZH\tnote{i}
& 100 talks (9.5 hours) 
& Abstractive (paper abstracts) 
& Manual
& \cmark 
& \cmark 
& CC-BY-4.0 \\  
\custommidrule
& \makebox[0pt][l]{\hypertarget{vista-row}{}}\href{https://github.com/dongqi-me/VISTA}{VISTA~\faExternalLink*} 
& \citet{liu_what_2025} 
& Conference talks (AI venues) 
& EN 
& 18.6k talks (2.1k hours) 
& Abstractive (paper abstracts) 
& \xmark
& \dmark\tnote{a} 
& \dmark\tnote{a} 
& \qmark\tnote{j} \\  
\custommidrule

& \makebox[0pt][l]{\hypertarget{slue-row}{}}\href{https://huggingface.co/datasets/asapp/slue-phase-2}{SLUE-TED~\faExternalLink*} 
& \citet{shon_slue_2023} 
& TED talks 
& EN 
& 4.2k talks (829 hours) 
& Abstractive (talk descriptions) 
& Manual 
& \cmark 
& \dmark\tnote{a} 
& CC\nobreakdash-BY\nobreakdash-NC-ND-4.0 \\
\custommidrule
\rmark\tnote{d} 
& \href{https://github.com/nttcslab-sp-admin/TEDSummary}{TEDSummary~\faExternalLink*} 
& \citet{kano_attention-based_2021} 
& TED talks 
& EN 
& 1.5k talks 
& Abstractive (talk descriptions) 
& Manual 
& \dmark\tnote{a} 
& \dmark\tnote{a} 
& \qmark\tnote{j} \\
\custommidrule
\rmark\tnote{e} 
& \href{https://github.com/GianlucaVico/ted-summarization}{TED Talk Teasers~\faExternalLink*} 
& \citet{vico_ted_2022} 
& TED talks 
& EN 
& 2.8k talks (739 hours)
& Abstractive (talk descriptions) 
& Manual 
& \dmark\tnote{a} 
& \dmark\tnote{a} 
& CC-BY-NC-ND-4.0 \\
\custommidrule
& \href{https://github.com/ucfnlp/streamhover}{StreamHover~\faExternalLink*} 
& \citet{cho_streamhover_2021} 
& Livestreams (Behance.net) 
& EN 
& 370 videos (500 hours) 
& \shortlong{Abs- \& Extractive (crowdsourced, clip-\& video-level)}{Abstractive \& Extractive (crowdsourced, clip-level \& video-level)}
& ASR 
& \dmark\tnote{a} 
& \dmark\tnote{a} 
& \qmark\tnote{j} \\
\seprule
& \href{https://github.com/zcgzcgzcg1/MediaSum}{MediaSum~\faExternalLink*} 
& \citet{zhu_mediasum_2021} 
& Media interviews (NPR, CNN) 
& EN 
& 463.6k interview segments 
& Abstractive (topic descriptions) 
& Manual 
& \xmark 
& \xmark 
& \qmark\tnote{j} \\
\custommidrule
& \makebox[0pt][l]{\hypertarget{summscreen-row}{}}\href{https://github.com/mingdachen/SummScreen}{SummScreen~\faExternalLink*} 
& \citet{chen_summscreen_2022} 
& TV show transcripts 
& EN 
& 26k episodes 
& Abstractive (episode recaps) 
& Manual 
& \xmark 
& \xmark 
& \qmark\tnote{j} \\
\seprule
\rmark\tnote{f} 
& \makebox[0pt][l]{\hypertarget{spotify-row}{}}\href{https://podcastsdataset.byspotify.com/}{Spotify Podcast Dataset~\faExternalLink*} 
& \citet{clifton_100000_2020,garmash_cem_2023} 
& Podcast episodes 
& EN, PT 
& 200k episodes (100k hours) 
& Abstractive (podcast descriptions) 
& ASR 
& \cmark 
& \xmark 
& \qmark\tnote{j} \\
\seprule
& \makebox[0pt][l]{\hypertarget{ami-row}{}}\href{https://groups.inf.ed.ac.uk/ami/corpus/}{AMI Meeting Corpus~\faExternalLink*} 
& \citet{carletta_ami_2006} 
& Business meetings (scenario-driven) 
& EN 
& 137 meetings (65 hours) 
& Abstractive \& Extractive (minutes), Topic segments 
& Manual 
& \cmark 
& \cmark 
& CC-BY-4.0 \\
\custommidrule
& \href{https://groups.inf.ed.ac.uk/ami/icsi/}{ICSI Meeting Corpus~\faExternalLink*} 
& \citet{janin_icsi_2003} 
& Research group meetings (naturalistic) 
& EN 
& 75 meetings (72 hours) 
& Abstractive \& Extractive (minutes), Topic segments 
& Manual 
& \cmark 
& \xmark 
& CC-BY-4.0 \\
\custommidrule
& \makebox[0pt][l]{\hypertarget{qmsum-row}{}}\href{https://github.com/Yale-LILY/QMSum}{QMSum~\faExternalLink*} 
& \citet{zhong_qmsum_2021} 
& AMI, ICSI \& Committee meetings 
& EN 
& 232 meetings 
& Abstractive (query-based, multiple), Topic segments 
& Manual 
& \xmark 
& \xmark 
& MIT \\
\custommidrule
&  \makebox[0pt][l]{\hypertarget{ELITR-row}{}}\href{https://ufal.mff.cuni.cz/elitr-minuting-corpus}{ELITR Minuting Corpus~\faExternalLink*} 
& \citet{nedoluzhko_elitr_2022} 
& Technical project \& parliament meetings (naturalistic) 
& EN, CS 
& 166 meetings (160 hours) 
& Abstractive (minutes, multiple) 
& Manual 
& \xmark 
& \xmark 
& CC-BY-NC-SA-4.0 \\
\custommidrule
& \href{https://huggingface.co/datasets/knkarthick/dialogsum}{DialogSum~\faExternalLink*} 
& \citet{chen_dialogsum_2021} 
& Diverse, spoken dialogues (EN-practicing scenarios) 
& EN 
& 13.4k dialogues 
& Abstractive (crowdsourced) 
& Manual 
& \xmark 
& \xmark 
& CC-BY-NC-SA-4.0 \\
\custommidrule
& \makebox[0pt][l]{\hypertarget{meetingbank-row}{}}\href{https://meetingbank.github.io/}{MeetingBank~\faExternalLink*} 
& \citet{hu_meetingbank_2023} 
& City council meetings (naturalistic) 
& EN 
& 1.3k meetings (3.5k hours) 
& Abstractive (segment-level minutes) 
& ASR 
& \cmark 
& \xmark 
& CC-BY-NC-ND-4.0 \\
\custommidrule
\incmark\tnote{g} 
& \href{https://github.com/ufal/europarlmin}{EuroParlMin~\faExternalLink*} 
& \citet{ghosal_overview_2023} 
& Parliament meetings (naturalistic) 
& EN 
& 2.2k sessions (1.8k hours)
& Abstractive (minutes) 
& Manual 
& \xmark 
& \xmark 
& \qmark\tnote{j} \\
\custommidrule
\incmark\tnote{g} 
& \href{https://mt.fbk.eu/europarl-interviews/}{EuroParl Interviews~\faExternalLink*} 
& \citet{papi-etal-2023-direct-speech} 
& Parliament meetings (naturalistic) 
& EN 
& 12 videos (1 hour) 
& Abstractive (sentence-level, cross-lingual) 
& Manual 
& \cmark 
& \cmark 
& CC-BY-NC-4.0 \\
\seprule
& \href{https://github.com/rajdeep345/ECTSum}{ECTSum~\faExternalLink*} 
& \citet{mukherjee_ectsum_2022} 
& Earnings calls (The Motley Fool)
& EN 
& 2.4k transcripts 
& Abstractive (bullet points, from Reuters) 
& Manual 
& \xmark 
& \xmark 
& GPL-3.0 \\
\seprule
& \href{https://huggingface.co/datasets/komats/mega-ssum}{MegaSSum~\faExternalLink*} 
& \citet{matsuura_sentence-wise_2024} 
& News articles (Gigaword, DUC2003) 
& EN 
& 3.8M articles 
& Abstractive (headlines) 
& N/A (Articles) 
& \smark\tnote{b} 
& \xmark 
& CC-BY-4.0 \\
\arrayrulecolor{black}\custombottomrule
\end{tabular}
\begin{tablenotes}[flushleft]\tiny\parindent=1em
    \setlength{\columnsep}{0.8cm}
    \setlength{\multicolsep}{0cm}

\setlength{\baselineskip}{6pt}
\shortlong{\setlength{\itemsep}{-2pt}}{}

    \begin{multicols}{2}
        \raggedcolumns
        \item[\text{\makebox[0.5em][l]{a}}] \makebox[0.8em][c]{\dmark} ~\strut Only a download script or source links are provided, but no direct data.
        \item[\text{\makebox[0.5em][l]{b}}] \makebox[0.8em][c]{\smark} ~\strut Data is synthesized rather than from real recordings.
        \item[\text{\makebox[0.5em][l]{c}}] \makebox[0.8em][c]{\rmark} ~\strut Unavailable since 12/2024 due to widespread video removals; no redistribution.
        \item[\text{\makebox[0.5em][l]{d}}] \makebox[0.8em][c]{\rmark} ~\strut Lacks documentation on included talks, hindering reproduction \cite{shon_slue_2023}.
        \item[\text{\makebox[0.5em][l]{e}}] \makebox[0.8em][c]{\rmark} ~\strut Reproduction hindered; lacking documentation and TED is no longer using Amara. 
        \item[\text{\makebox[0.5em][l]{f}}] \makebox[0.8em][c]{\rmark} ~\strut Unavailable since 12/2023 due to resource constraints.
        \item[\text{\makebox[0.5em][l]{g}}] \makebox[0.8em][c]{\incmark} ~\strut Not all data partitions are available (only test set or no test set).
        \item[\text{\makebox[0.5em][l]{h}}] \makebox[0.8em][c]{\incmark} ~\strut  Partial language availability (only transcript translations).
        \item[\text{\makebox[0.5em][l]{i}}] \makebox[0.8em][c]{\incmark} ~\strut  Partial language availability (only summary translations).
        \item[\text{\makebox[0.5em][l]{j}}] \makebox[0.8em][c]{\qmark} ~\strut No explicit license has been provided.
            \end{multicols}
\end{tablenotes}
\end{threeparttable}
\caption{English and multilingual datasets related to the SSum task. Datasets that are exclusively non-English, chat-based datasets, and derivatives or extensions of existing resources are listed in Tables~\ref{tab:nonenglish-datasets}, \ref{tab:chat-datasets}, and~\ref{tab:dataset-derivatives}.}
\label{tab:datasets}\shortlong{\vspace{-0.245cm}}{}
\end{table*}

%% file: 05_eval.tex
\shortlong{
\section{Evaluation of Speech Summaries}
\label{sec:eval}
Accurately evaluating SSum systems is crucial for measuring progress and ensuring reliable outputs, yet it remains challenging. First, there is no single ground truth for summaries, as humans emphasize different aspects and phrase information variably  \citep{rathdifferentgroundtruths1961, harman-over-2004-effects, clark-etal-2021-thats, cohan-etal-2022-overview-first, Sharma2024, zhang-etal-2024-benchmarking}. This is especially true for speech summaries such as podcast summaries, which tend to be longer and more abstractive \citep{manakul_podcast_2022} compared to domains like news summarization.  Moreover, summaries often differ when based on transcripts versus audio \citep{sharma-etal-2024-speech}. Second, evaluators struggle with summaries as their length and varied wording make evaluation difficult \citep{goyal2023newssummarizationevaluationera}.
Lastly, evaluating quality requires assessing lexical, semantic, and factual correctness \citep{liu-etal-2023-g, Kroll_2024, Sharma2024}, which makes the evaluation process complex. Even with reference comparisons, human evaluations are often inconsistent \citep{hardy-etal-2019-highres}.

While TSum evaluation already presents challenges, SSum adds further complexity due to the characteristics of spoken language. \citet{kirstein-etal-2024-whats} show that colloquialisms, background noise, and multiple speakers introduce unique errors, such as speaker misidentification affecting pronoun usage \citep{rennard_abstractive_2023}. Additionally,  cascaded models further propagate transcription errors into summarization  \citep{zechner_minimizing_2000, rennard_abstractive_2023, chowdhurytranscripterrors2024} and its evaluation \citep{sharma-etal-2024-speech}.

SSum evaluation methods range from human assessments to automated metrics, including lexical overlap like ROUGE \citep{lin-2004-rouge}, embedding-based metrics such as BERTScore \citep{zhang_bertscore_2020}, and model-based evaluators like fact-checking systems or LLM judges. However, popular SSum evaluation methods, like ROUGE and BERTScore, remain grounded in TSum approaches and often overlook the distinct challenges posed by spoken content. For example, the unstructured nature of \shortlong{}{spontaneous} speech reduces ROUGE’s correlation with human judgment \citep{liu-liu-2008-correlation}, while \citet{kirstein_whats_2024} find that BERTScore has not been thoroughly evaluated for meeting summarization and is often unsuitable due to its \shortlong{}{512-token} context limit, frequently exceeded by lengthy transcripts.

In the following sections, we focus on human and LLM-as-a-Judge evaluation, as these can be better tailored to 
SSum. Other metrics originally developed for TSum are reviewed in detail in \cref{app:eval}. \cref{fig:metrics_over_years}  illustrates the use of these metrics over time, highlighting the growing popularity of LLM-based and trained evaluator metrics compared to traditional lexical overlap metrics.

\subsection{Human Evaluation}

Human evaluation is often considered the gold standard for assessing summarization quality \citep{clark-etal-2021-thats} and enables assessment of specific speech-related content. For example, in podcast SSum, details like episode structure and host-guest roles can be evaluated, reflecting the unique nature of spoken media \citep{song_towards_2022}. In meeting summarization, other evaluations have focused on how well summaries capture decision-making content from the meeting \citep{murray2009decisionaudio}.

However, human annotation presents several challenges: it requires extensive effort \citep{card-etal-2020-little} and is both time-consuming and costly. This is especially true for long meeting summaries, where annotators must watch lengthy videos, read full transcripts, and evaluate each system-generated summary based on multiple criteria \citep{hu_meetingbank_2023}. ASR errors in the transcript might make this process even more challenging \citep{murray2009decisionaudio}. Moreover, the lack of a standardized procedure—despite several proposed frameworks \citep{nenkova-passonneau-2004-evaluating, hardy-etal-2019-highres, liu-etal-2023-revisiting, Kroll_2024}—further complicates large-scale assessments \citep{iskender_best_2020}.

High-quality evaluations often depend on costly expert judgments \citep{gillick-liu-2010-non}. In SSum, the length and complexity of transcripts or even full audio recordings further increase the effort required. Crowdsourcing offers a more affordable alternative, and with appropriate guidelines, crowd workers can achieve expert-level performance \citep{iskender_best_2020, iskender-etal-2020-towards}. However, such evaluations tend to be more uniform and often struggle with identifying nuanced errors \citep{fabbri_summeval_2021}.

Evaluations may be conducted either referenceless \citep{song_towards_2022, goyal2023newssummarizationevaluationera, schneider2025policiesevaluationonlinemeeting} or with references \citep{fabbri_summeval_2021, zufle_nutshell_2025}, but these setups often show low inter-method correlation \citep{liu-etal-2023-revisiting}, making results difficult to compare. 

A detailed overview of human evaluation protocols for SSum is provided in \cref{tab:human_eval}. Notably, most human evaluations rely solely on transcripts, which simplifies the process but neglects important auditory cues such as prosody, pauses, and speaker dynamics. Indeed, previous work has shown that speech-based summaries tend to be more factually consistent and information-selective than transcript-based summaries \citep{sharma-etal-2024-speech}.

\subsection{LLM-as-a-Judge}

Using LLMs as evaluators is an emerging approach where models are prompted to assess summaries directly \citep{shen-etal-2023-large, liu-etal-2023-g, zheng-llm-judge-2024, gong_cream_2024, kirstein_is_2025}. These models are applied by calculating win rates against reference models \citep{dubois_alpacafarm_2023, dubois_length-controlled_2024}, evaluating specific criteria \citep{liu-etal-2023-g, tang_tofueval_2024, zufle_nutshell_2025}, and performing reference-free quality estimation \citep{liu-etal-2023-g, gong_cream_2024, kirstein_is_2025}.  \cref{tab:llm_judge} shows an overview of these approaches. Among these, CREAM \citep{gong_cream_2024}, MESA \citep{kirstein_is_2025}, and TofuEval \citep{ tang_tofueval_2024} stand out as one of the few frameworks specifically developed for meeting and dialogue summarization, targeting long-context summarizations and dialogue-based meeting summarizations. Notably, the LLM-based evaluators either rely on transcripts or use only the system output and reference summaries to reduce computational costs. To date, no models evaluate the SSum content directly from raw audio signals.

Still, LLM judges show strong performance, often surpassing traditional metrics like ROUGE and aligning closer with human judgments \citep{zufle_nutshell_2025}. However, it comes with limitations: The judge must be stronger than the systems it assesses \citep{dubois_alpacafarm_2023}, often involving commercial models with limited reproducibility \citep{barnes2025summarizationmetricsspanishbasque}. LLM judges also exhibit biases, such as favoring outputs from the same model \citep{dubois_alpacafarm_2023, gong_cream_2024}, struggling with factual error detection \citep{gong_cream_2024, tang_tofueval_2024}, preferring list-style over fluent text \citep{dubois_alpacafarm_2023}, and being sensitive to prompt complexity \citep{thakur2025judgingjudgesevaluatingalignment} and summary length \citep{dubois_length-controlled_2024,thakur2025judgingjudgesevaluatingalignment}. 
These limitations are particularly relevant for SSum, where current LLM-based evaluators do not process audio or even the transcript, failing to account for key characteristics of speech such as prosody.

}{
\section{Evaluation of Speech Summaries}
\label{sec:eval}

Accurately evaluating SSum systems is crucial for measuring progress and ensuring reliable and faithful outputs, yet it remains challenging. First, there is no definitive ground truth for summaries, as humans emphasize different aspects and phrase information variably  \citep{rathdifferentgroundtruths1961, harman-over-2004-effects, clark-etal-2021-thats, cohan-etal-2022-overview-first, Sharma2024, zhang-etal-2024-benchmarking}. This is especially true for speech summaries such as podcast summaries, which tend to be longer and more abstractive \citep{manakul_podcast_2022} compared to domains like news summarization.  Moreover, summaries often differ when based on transcripts versus audio \citep{sharma-etal-2024-speech}. Second, evaluators struggle with multi-sentence summaries as their length and varied wording make evaluation difficult \citep{goyal2023newssummarizationevaluationera, mastropaolo2023evaluatingcodesummarizationtechniques}. Lastly, evaluating quality requires assessing lexical, semantic, and factual correctness \citep{liu-etal-2023-g, Kroll_2024, Sharma2024}, which makes the evaluation process complex. Even with reference comparisons, human evaluations are often inconsistent \citep{hardy-etal-2019-highres}. 

While TSum evaluation already presents challenges, evaluating SSum introduces additional complexities due to the characteristics of spoken language. \citet{kirstein-etal-2024-whats} show that colloquialisms, background noise, and multiple speakers introduce unique errors to the summaries, such as speaker misidentification affecting pronoun usage \citep{rennard_abstractive_2023}. Cascaded models further propagate transcription errors into summarization  \citep{zechner_minimizing_2000, rennard_abstractive_2023, chowdhurytranscripterrors2024}. However, current evaluation methods for SSum remain grounded in TSum approaches, which may overlook the distinct challenges of spoken content. For instance, they often fail to account for speaker attribution errors or relevant background noise that impact the coherence and accuracy of the summary.

Evaluation methods for SSum range from human assessments to automated metrics. These include lexical overlap metrics—most notably ROUGE \citep{lin-2004-rouge}—embedding-based metrics such as BERTScore \citep{zhang_bertscore_2020}, and model-based evaluators, for example, fact-checking systems or LLMs used as judges. 
\cref{fig:metrics_over_years} illustrates the use of these metrics over time, highlighting the growing popularity of LLM-based and trained evaluator metrics compared to traditional lexical overlap and embedding-based metrics. 
In the following, we discuss these metrics in detail.

\subsection{Human Evaluation.}
Human evaluation is often considered the gold standard for assessing summarization quality \citep{clark-etal-2021-thats} and enables assessment of specific speech-related content. For example, in podcast SSum, details like the episode structure of podcasts and roles of hosts and guests can be evaluated, reflecting the unique nature of spoken media \citep{song_towards_2022}. In meeting summarization, other evaluations have focused on specific aspects such as how well summaries capture decision-making content from the meeting \citep{murray2009decisionaudio}.

However, human annotation presents several challenges: it requires extensive effort \citep{card-etal-2020-little} and is both time-consuming and costly. This is especially true for long meeting summaries, where annotators must watch lengthy videos, read full transcripts, and evaluate each system-generated summary based on multiple criteria \citep{hu_meetingbank_2023}. ASR errors in the transcript might make this process even more challenging \citep{murray2009decisionaudio}. Moreover, the lack of a standardized procedure—despite several proposed frameworks \citep{nenkova-passonneau-2004-evaluating, hardy-etal-2019-highres, liu-etal-2023-revisiting, Kroll_2024}—further complicates large-scale assessments \citep{iskender_best_2020}.

In addition, high-quality evaluations often depend on costly expert judgments \citep{gillick-liu-2010-non}. These challenges are particularly pronounced in SSum, where longer and more complex transcripts—or even full audio recordings—further increase the time and cost of manual evaluation. Crowdsourcing offers a more affordable alternative, and with appropriate guidelines, crowd workers can achieve expert-level performance \citep{iskender_best_2020, iskender-etal-2020-towards}. However, such evaluations tend to be more uniform and often struggle with identifying nuanced errors \citep{fabbri_summeval_2021}.

Evaluations may be conducted either referenceless \citep{song_towards_2022, goyal2023newssummarizationevaluationera, schneider2025policiesevaluationonlinemeeting} or with references \citep{fabbri_summeval_2021, zufle_nutshell_2025}, but these setups often show low inter-method correlation \citep{liu-etal-2023-revisiting}, making results difficult to compare. 

A detailed overview of different human evaluation protocols for SSum is provided in \cref{tab:human_eval}. Notably, most human evaluations for SSum rely solely on transcripts, which simplifies the process but neglects important auditory cues such as intonation, pauses, and speaker dynamics. Indeed, previous work has shown that speech-based summaries tend to be more factually consistent and information-selective than transcript-based summaries \citep{sharma-etal-2024-speech}.

\subsection{Lexical Overlap Metrics.}
Lexical overlap metrics assess similarity based on shared surface-level units. ROUGE \citep{lin-2004-rouge}, designed to maximize recall, is the most widely used metric \citep{fabbri_summeval_2021, Sharma2024}, though implementation errors have led to incorrect evaluations in the past \citep{grusky-2023-rogue}. 
Moreover, early work has shown that the presence of disfluencies, multiple speakers, and the lack of structure in spontaneous speech diminish the correlation between ROUGE scores and human judgment \citep{liu-liu-2008-correlation}.
BLEU \citep{papineni-etal-2002-bleu, post-2018-call} and METEOR \citep{banerjee-lavie-2005-meteor} remain common to evaluate summaries despite being developed for machine translation.
Methods like Basic Elements \citep{hovy-etal-2006-automated} and the Pyramid Method \citep{nenkova-passonneau-2004-evaluating} improve overlap metrics by also considering syntactic dependencies and content units.

Despite their efficiency, these lexical overlap metrics struggle to evaluate faithfulness to the input \citep{bhandari-etal-2020-metrics, maynez-etal-2020-faithfulness, wang-etal-2020-asking}, fail to distinguish similar or high-scoring candidates \citep{peyrard-2019-studying, bhandari-etal-2020-metrics}, and are often outperformed by model-based evaluators, which has been shown for dialog summarization by \citet{gao-wan-2022-dialsummeval}. Since they do not use the source speech or transcript, they often fail to account for SSum-specific attributes.

\subsection{Model-Based Evaluators}
\paragraph{Embedding-Based Metrics.}
Embedding-based metrics capture semantic similarity through sentence or token embeddings. Yet, they still struggle to assess factual accuracy, fully capture shared information \citep{deutsch-roth-2021-understanding}, and distinguish similar candidates \citep{bhandari-etal-2020-metrics}. 

BERTScore \citep{zhang_bertscore_2020}, one of the most prominent embedding-based metrics, compares contextualized token embeddings between the summary and reference. Yet, \citet{kirstein_whats_2024} find that BERTScore has not been tested for meeting summarization and is often unsuitable due to its 512-token context limit, which is frequently exceeded by lengthy transcripts. Other model-based evaluators include MoverScore \citep{zhao-etal-2019-moverscore}, which measures the earth-mover distance between embeddings, capturing both content overlap and divergence, and SPEEDScore \citep{akula_sentence_2022}, which evaluates summary efficiency by balancing compression and information retention using sentence-level embeddings.

\paragraph{Trained Evaluators.}
Recent approaches have focused on training models for more holistic summary evaluation \citep{NEURIPS2021_e4d2b6e6, zhong_towards_2022}, as well as for specific dimensions like factual accuracy. The latter can  be evaluated by defining a question-answering based metric such as FEQA \citep{durmus-etal-2020-feqa}, QAGS \citep{wang-etal-2020-asking} or QuestEval \citep{scialom-etal-2021-questeval}, or by explicitly training a model for this task \citep{kryscinski_evaluating_2020}.
Other models refine evaluations using counterfactual estimation \citep{xie-etal-2021-factual-consistency} and causal graphs \citep{ling-etal-2025-enhancing}. However, even evaluation-specific models, particularly reference-free ones, may be prone to spurious correlations such as summary length \citep{durmus_spurious_2022}.

\paragraph{LLM-as-a-Judge.}
Using LLMs as evaluators is an emerging approach where models are prompted to assess summaries directly \citep{shen-etal-2023-large, liu-etal-2023-g, zheng-llm-judge-2024, gong_cream_2024, kirstein_is_2025}. These models are applied by calculating win rates against reference models \citep{dubois_alpacafarm_2023, dubois_length-controlled_2024}, evaluating specific criteria \citep{liu-etal-2023-g, tang_tofueval_2024, zufle_nutshell_2025}, and performing reference-free quality estimation \citep{liu-etal-2023-g, gong_cream_2024, kirstein_is_2025}.  \cref{tab:llm_judge} shows an overview of these approaches. Among these, CREAM \citep{gong_cream_2024}, MESA \citep{kirstein_is_2025}, and TofuEval \citep{ tang_tofueval_2024} stand out as one of the few frameworks specifically developed for meeting and dialogue summarization, targeting long-context summarizations and dialogue-based meeting summarizations. Notably, the LLM-based evaluators either rely on transcripts or use only the system output and reference summaries to reduce computational costs. To date, no models evaluate the SSum content directly from raw audio signals.

Still, LLM-as-a-Judge has shown strong performance, often surpassing traditional metrics like ROUGE and aligning closer with human judgments \citep{zufle_nutshell_2025}. However, they come with limitations: The judge model must be stronger than the systems it assesses \citep{dubois_alpacafarm_2023}, often involving commercial models with limited reproducibility \citep{barnes2025summarizationmetricsspanishbasque}. LLM judges also exhibit biases, such as favoring outputs from the same model \citep{dubois_alpacafarm_2023, gong_cream_2024}, struggling with factual error detection \citep{gong_cream_2024, tang_tofueval_2024}, preferring list-style over fluent text \citep{dubois_alpacafarm_2023}, and being sensitive to prompt complexity \citep{thakur2025judgingjudgesevaluatingalignment} and summary length \citep{dubois_length-controlled_2024,thakur2025judgingjudgesevaluatingalignment}. They also have difficulty distinguishing similar candidates \citep{shen-etal-2023-large} and suffer from position bias, where earlier outputs receive higher scores \citep{wang-etal-2024-large-language-models-fair, dubois_alpacafarm_2023}.  

Some of these issues can be mitigated by controlling for length biases and predicting evaluator preferences \citep{dubois_length-controlled_2024}, or using Set-LLM to avoid position bias \citep{egressy2025setllmpermutationinvariantllm}. However, the biggest flaw remains, namely that current LLM-based evaluators do not process audio or even the transcript, and hence fail to account for key characteristics of speech such as its prosodically rich and multi-speaker nature.
}

%% file: 06_approaches.tex
\section{Approaches}\label{sec:approaches}
\subsection{Cascaded Approaches}

Cascaded approaches remain the most widely adopted paradigm in SSum. In this framework, speech is first transcribed using an ASR system and then passed to a TSum model. Two primary strategies have emerged in this paradigm: first, fine-tuning of ED models specifically for summarization, and second, prompting and adapting LLMs.

\subsubsection{Fine-Tuning Encoder-Decoder Models}

To enable cascaded approaches for SSum, many works focused on fine-tuning pretrained ED models such as BART, Longformer/LED, PEGASUS, DialogLM, and HMNet \cite[e.g.,][]{zhong_qmsum_2021,hu_meetingbank_2023,huang_swing_2023,fu_tiny_2024,le-duc_real-time_2024,zhu_factual_2025}, ranging from general-purpose models such as BART and Longformer/LED to more specialized models. PEGASUS \cite{zhang_pegasus_2020}, for example, incorporates a summarization-specific pretraining using \textit{gap sentences generation} while DialogLM/DialogLED \cite{zhong_dialoglm_2022} is trained on denoising with dialogue-inspired noise.

\paragraph{Handling Long Context.} Long input is a particular concern for SSum, as spoken content often yields lengthy, unstructured transcripts with dispersed information. As such, many works rely on Longformer \cite{beltagy_longformer_2020} or explore alternative sparse or windowed attention mechanisms \cite{zhang_exploratory_2021,zhong_dialoglm_2022}. Alternatively, researchers have explored hierarchical encoders \cite{zhu_hierarchical_2020,zhang_exploratory_2021}, retrieve-then-summarize or locate-then-summarize strategies \cite{zhang_exploratory_2021,zhong_qmsum_2021}, and segment-level processing \cite{zhang_summn_2022,laskar_building_2023,retkowski_text_2024}.

\paragraph{Robustness and Faithfulness.} Faithfulness is a central challenge in summarization and is particularly problematic in cascaded SSum due to ASR error propagation. To improve robustness, some approaches fuse multiple ASR hypotheses \cite{xie_using_2010,kano_attention-based_2021} or ground summary segments to the transcript \cite{song_towards_2022}. To enhance faithfulness, other works apply symbolic knowledge distillation \cite{zhu_factual_2025} or incorporate fine-grained entailment signals during training \cite{huang_swing_2023,kim_explainmeetsum_2023}.

\shortlong{\input{tables/llm_approaches}}{}

\paragraph{Contextual and Multimodal Enrichment.} Some approaches enrich SSum models with additional contextual or multimodal signals, such as speaker-role information \cite{zhu_hierarchical_2020}, video features combined with transcripts \cite{palaskar_multimodal_2019}, or joint representations of text, video, and speech concepts \cite{palaskar_multimodal_2021}.


\subsubsection{Prompting and Adapting LLMs}

More recently, LLMs have enabled zero-shot SSum through prompting without the need for task-specific training. This capability has been explored on various models such as GPT-3.5, PaLM-2, and LLaMA 3 \cite{hu_meetingbank_2023,fu_tiny_2024,nelson_cross-lingual_2024,zufle_nutshell_2025}. Building on this, several studies propose more sophisticated prompting strategies, including few-shot prompting and iterative self-refinement \cite{laskar_building_2023,kirstein-etal-2024-whats}. To improve performance and efficiency, methods such as LoRA fine-tuning for SSum-specific adaptation \cite{nelson_cross-lingual_2024} and knowledge distillation into smaller models \cite{fu_tiny_2024,zhu_factual_2025} have been applied.

\subsection{End-to-End Approaches}
\label{sec:end2end}

E2E SSum has recently gained significant traction as a research area, with models that directly map raw audio to textual summaries without relying on an intermediate transcription. They fall broadly into two categories: task-specific architectures designed and trained directly for SSum, and modular systems that integrate LLMs with audio encoders via projection mechanisms.

\shortlong{}{\input{tables/llm_approaches}}

\subsubsection{Task-Specific Models}
These models often follow a two-stage training paradigm: first, a pretraining on ASR tasks to learn the mapping from speech to text and to acquire rich acoustic-linguistic representations, followed by summarization fine-tuning \citep[e.g.,][]{chen_train_2024,eom_squba_2025}. However, in contrast to other speech-processing tasks like ASR, SSum effectively demands the full context of the document. This poses a challenge for the original Transformer architecture, whose self-attention mechanism scales quadratically with input length, making it inefficient for long sequences.  To overcome this, researchers typically rely on input speech truncation \cite{matsuura_leveraging_2023,sharma_bass_2023,chen_train_2024} or input compression such as temporal downsampling \cite{chu_qwen2-audio_2024,kang_prompting_2024} or higher-level/segment-level projections \cite{shang_end--end_2024}. Others have explored more fundamental architectural modifications, including adjusting the attention mechanism \cite{sharma_end--end_2022,sharma_bass_2023,sharma_r-bass_2024} or replacing it entirely with more efficient structures such as FNet \cite{kano_speech_2023,chen_train_2024}, convolutions \cite{chen_train_2024}, or state-space models like Mamba \cite{miyazaki_exploring_2024,eom_squba_2025}.


\subsubsection{LLM-Based Systems}
In parallel, efforts to leverage pretrained language models have gained momentum: earlier work explored transfer learning from ED models like BART \cite{matsuura_transfer_2023}, while more recent approaches focus on directly integrating pretrained LLMs by attaching an \textit{audio encoder}. As shown in \cref{tab:audio-llm-approaches}, these methods typically pair an audio encoder---such as Conformer \cite{fathullah_audiochatllama_2024,shang_end--end_2024,microsoft_phi-4-mini_2025}, HuBERT \cite{kang_prompting_2024,zufle_nutshell_2025}, or Whisper \cite{chu_qwen2-audio_2024,eom_squba_2025, he_meralion-audiollm_2025}---with a \textit{projection module} such as a Q-Former \cite{shang_end--end_2024,zufle_nutshell_2025}, MLP \cite{he_meralion-audiollm_2025,microsoft_phi-4-mini_2025}, or linear layer \cite{chu_qwen2-audio_2024,fathullah_audiochatllama_2024,kang_prompting_2024} that maps audio features into the LLM's input space. These configurations differ in how much or which part of the system is trained. While all approaches train a projection module, they vary in whether they also fine-tune the audio encoder or the LLM. Some methods keep both components frozen, training only the projector \cite{zufle_nutshell_2025}. Others \cite{fathullah_audiochatllama_2024, kang_prompting_2024, microsoft_phi-4-mini_2025} train the projector alongside the audio encoder. Several approaches fine-tune the LLM using parameter-efficient techniques such as LoRA \cite{shang_end--end_2024, he_meralion-audiollm_2025}. \citet{chu_qwen2-audio_2024} instead adopt full end-to-end training, keeping all parameters of the audio encoder, projector, and LLM trainable. \citet{eom_squba_2025} propose an alternative to transformer-based systems using Q-Mamba and a pretrained Mamba LLM.

\paragraph{Zero-Shot E2E SSum.}

LLM-based open-source models now, for the first time, make E2E SSum accessible with minimal setup. Models like Qwen2-Audio \cite{chu_qwen2-audio_2024} have been used for zero-shot SSum without task-specific training \cite{he_meralion-audiollm_2025,zufle_nutshell_2025}. Similarly, Phi-4 \cite{microsoft_phi-4-mini_2025} supports audio inputs and shows potential for general-purpose SSum.


\shortlong{}{\input{tables/quanitative_synthesis}}

\subsection{Quantitative Synthesis}
\Cref{tab:quantitative} synthesizes reported scores on \hyperlink{how2-row}{How2} across end-to-end systems and their cascaded baselines. Due to the diverse landscape of evaluation protocols and benchmarks in SSum, only end-to-end approaches could be compared \shortlong{meaningfully}{in a meaningful way}, and only on How2, using ROUGE and BERTScore as evaluation metrics. Within this scope, E2E models generally outperform cascaded approaches, with performance shaped \shortlong{}{most strongly} by the amount of context a model can process, its parameter count, and whether input data is enriched with synthetic speech. Systems handling longer or full inputs surpass those limited to truncated segments, underscoring the importance of long-context handling and the potential of alternative architectures.

%% file: tables/llm_approaches.tex
\begin{table*}[!htb]
\centering
\renewcommand{\arraystretch}{1.0}
\setlength{\tabcolsep}{0.6em}
\tiny
\newcommand{\customtoprule}{\specialrule{1pt}{0pt}{2pt}}
\newcommand{\custombottomrule}{\specialrule{1pt}{2pt}{0pt}}
\newcommand{\custommidrule}{\specialrule{0.3pt}{1.5pt}{1.5pt}}
\begin{threeparttable}
\begin{tabular}{P{2.0cm} P{3.6cm} P{2.8cm} P{4.6cm}}
\customtoprule
\textbf{Reference} & \textbf{Audio Encoder} & \textbf{Projector} & \textbf{LLM} \\
\custommidrule

\citet{fathullah_audiochatllama_2024}
& \firemark\ Conformer \cite{gulati_conformer_2020} 
& \firemark~Linear 
& \snowmark\ LLaMA-2-7B-chat \cite{touvron_llama_2023} \\

\custommidrule

\citet{shang_end--end_2024}
& \firemark\ Conformer \cite{gulati_conformer_2020} 
& \firemark\ Q-Former \cite{li_blip-2_2023} 
& \smark\ LLaMA-2-7B-chat \cite{touvron_llama_2023} \\

\custommidrule

\citet{microsoft_phi-4-mini_2025}
& \firemark\ Conformer \cite{gulati_conformer_2020} 
& \firemark~MLP
& \snowmark\ Phi-4-mini-instruct \cite{microsoft_phi-4-mini_2025} \\

\custommidrule

\citet{kang_prompting_2024}
& \firemark\ HuBERT-Large \cite{hsu_hubert_2021} 
& \firemark~Linear 
& \snowmark\ MiniChat-3B \cite{zhang_towards_2024} \\

\custommidrule

\citet{zufle_nutshell_2025}
& \snowmark\ HuBERT-Large \cite{hsu_hubert_2021} 
& \firemark\ Q-Former \cite{li_blip-2_2023} 
& \snowmark\ LLaMA3.1-8B-Instruct \cite{grattafiori2024llama3herdmodels} \\

\custommidrule

\citet{he_meralion-audiollm_2025}
& \snowmark\ MERaLiON-Whisper \cite{he_meralion-audiollm_2025} 
& \firemark~MLP
& \smark\ SEA-LION V3 \cite{he_meralion-audiollm_2025} \\

\custommidrule

\citet{chu_qwen2-audio_2024}
& \firemark\ Whisper-large-v3 \cite{radford_robust_2023} 
& \firemark~Linear 
& \firemark\ Qwen-7B \cite{bai_qwen_2023} \\

\custommidrule

\citet{eom_squba_2025}
& \snowmark\ Whisper-large-v2 \cite{radford_robust_2023} 
& \firemark\ Q-Mamba \cite{eom_squba_2025} 
& \firemark~Mamba-2.8B-Zephyr (\href{https://hf.co/xiuyul/mamba-2.8b-zephyr}{xiuyul/mamba-2.8b-zephyr}) \\

\custombottomrule
\end{tabular}
\end{threeparttable}
\caption{Overview of Audio Encoder $\rightarrow$ Projector $\rightarrow$ LLM Architectures (\firemark~trainable, \snowmark~frozen, \smark~LoRA)\shortlong{\vspace{-0.3cm}}{}}
\label{tab:audio-llm-approaches}
\end{table*}

%% file: tables/quanitative_synthesis.tex
\begin{table*}[!htb]
\centering
\providecommand{\customtoprule}{\specialrule{1pt}{0pt}{2pt}}
\providecommand{\custombottomrule}{\specialrule{1pt}{2pt}{0pt}}
\providecommand{\custommidrule}{\specialrule{0.3pt}{1.5pt}{1.5pt}}
\renewcommand{\arraystretch}{1.0}
\setlength{\tabcolsep}{0.55em}
\small

\resizebox{0.93\textwidth}{!}{%

\begin{threeparttable}

\begin{tabular}{!{\color{white}\vrule width 2pt}
                P{4.35cm} P{0.5cm} P{1.4cm} P{2.9cm} P{0.75cm} P{0.75cm} P{3.2cm}}
\customtoprule
\textbf{Architecture} & \textbf{Year} & \textbf{Input} & \textbf{\# Params} & \textbf{R-L} & \textbf{BS} & \textbf{Reported By} \\
\end{tabular}

\begin{tabular}{!{\color{ctxorange!30}\vrule width 2pt}
                P{4.35cm} P{0.5cm} P{1.4cm} P{2.9cm} P{0.75cm} P{0.75cm} P{3.2cm}}
\custommidrule
\rowcolor{ctxorange!10}\multicolumn{7}{l}{\textbf{\textsc{Cascaded systems}}}\\[-0.5ex]
\custommidrule
Conformer + BART-base      & 2022 & \inputtrunc{100} & 107M+140M   & \cellcolor[HTML]{ffffff} 50.3 & \cellcolor[HTML]{ffffff} 90.3 & \citet{sharma_end--end_2022} \\
Conformer + BART-large     & 2022 & \inputtrunc{100} & 107M+400M   & \cellcolor[HTML]{ffffff} 52.3 & \cellcolor[HTML]{ffffff} 90.6 & \citet{sharma_end--end_2022} \\
Conformer + BART-base      & 2023 & \inputtrunc{100} & 201M+140M   & \cellcolor[HTML]{ffffff} 55.4 & \cellcolor[HTML]{e9f5da} 92.6 & \citet{matsuura_transfer_2023} \\
Whisper-base + T5 Base     & 2023 & \inputfull       & 74M+220M    & \cellcolor[HTML]{f5faee} 57.5 & \cellcolor[HTML]{ffffff} 91.5 & \citet{sharma_espnet-summ_2023} \\
Conformer + LLaMA 2 7B \zeroshotmark     & 2024 & \inputfull       & \string~200M
 +7B / \string~200M\tnote{\firemark} & \cellcolor[HTML]{f0f8e6} 58.6 & \cellcolor[HTML]{e7f4d5} 91.8 & \citet{shang_end--end_2024} \\
\end{tabular}

\begin{tabular}{!{\color{ctxteal!30}\vrule width 2pt}
                P{4.35cm} P{0.5cm} P{1.4cm} P{2.9cm} P{0.75cm} P{0.75cm} P{3.0cm}}
\customtoprule
\rowcolor{ctxteal!10}\multicolumn{7}{l}{\textbf{\textsc{End-to-end models}}}\\[-0.5ex]
\custommidrule
Longformer-Transformer (RSA) & 2022 & \inputtrunc{100} & 104M & \cellcolor[HTML]{f9fdf8} 56.1 & \cellcolor[HTML]{ffffff} 91.5 & \citet{sharma_end--end_2022} \\
Whisper-base (Fine-Tuned)            & 2023 & \inputtrunc{30}  & 74M  & \cellcolor[HTML]{ffffff} 54.4 & \cellcolor[HTML]{ffffff} 88.5 & \citet{sharma_espnet-summ_2023} \\
Conformer-Transformer        & 2023 & \inputtrunc{30}  & 203M & \cellcolor[HTML]{e4f2ce} 59.2 & \cellcolor[HTML]{e4f2ce} 92.1 & \citet{sharma_espnet-summ_2023} \\
Conformer-Transformer (BASS)        & 2023 & \inputtrunc{30}  & 103M & \cellcolor[HTML]{dcf0c2} 60.2 & \cellcolor[HTML]{e1f1ca} 92.5 & \citet{sharma_bass_2023} \\
Conformer-Transformer        & 2024 & \inputtrunc{100} & 98M  & \cellcolor[HTML]{dbefc0} 60.5 & \cellcolor[HTML]{e1f1ca} 92.5 & \citet{miyazaki_exploring_2024} \\
Mamba-Transformer            & 2024 & \inputtrunc{100} & 96M  & \cellcolor[HTML]{ccebac} 62.3 & \cellcolor[HTML]{d7edb7} 92.9 & \citet{miyazaki_exploring_2024} \\
Mamba-Transformer            & 2024 & \inputtrunc{600} & 96M  & \cellcolor[HTML]{c9eaa6} 62.9 & \cellcolor[HTML]{d5edb4} 93.1 & \citet{miyazaki_exploring_2024} \\
Conformer-Transformer        & 2023 & \inputtrunc{100} & 203M & \cellcolor[HTML]{d0ecad} 62.0 & \cellcolor[HTML]{d5edb4} 93.2 & \citet{matsuura_leveraging_2023} \\
Conformer-Transformer \augmark & 2023 & \inputtrunc{100} & 203M & \cellcolor[HTML]{b4e179} 65.0 & \cellcolor[HTML]{b4e179} 93.8 & \citet{matsuura_leveraging_2023} \\
Conformer-BART-base \augmark & 2023 & \inputtrunc{100} & 203M & \cellcolor[HTML]{bce488} 63.2 & \cellcolor[HTML]{b4e179} 94.0 & \citet{matsuura_transfer_2023} \\
FNet-Transformer                          & 2024 & \inputfull       & 82M  & \cellcolor[HTML]{bae385} 63.6 & \cellcolor[HTML]{b8e281} 93.7 & \citet{chen_train_2024} \\
Conv-Transformer                   & 2024 & \inputfull       & 82M  & \cellcolor[HTML]{b8e281} 64.1 & \cellcolor[HTML]{b7e280} 93.6 & \citet{chen_train_2024} \\
Conformer-Transformer (Flash) & 2024 & \inputfull       & 95M  & \cellcolor[HTML]{b4e179} 65.5 & \cellcolor[HTML]{b4e179} 93.9 & \citet{chen_train_2024} \\
Conformer + QF + LLaMA 2 7B     & 2024 & \inputfull       & 7.2B / \string~215M\tnote{\firemark} & \cellcolor[HTML]{e1f1ca} 59.7 & \cellcolor[HTML]{b4e179} 93.9 & \citet{shang_end--end_2024} \\
Conformer + QF + LLaMA 2 13B     & 2024 & \inputfull       & 13.2B  / \string~220M\tnote{\firemark} & \cellcolor[HTML]{e2f2cb} 59.4 & \cellcolor[HTML]{b4e179} 93.9 & \citet{shang_end--end_2024} \\

\custombottomrule
\end{tabular}

\begin{tablenotes}[flushleft]\small\parindent=1em
    \setlength{\columnsep}{0.8cm}
    \setlength{\multicolsep}{0cm}

\setlength{\baselineskip}{12pt}
    
    \begin{multicols}{2}
        \raggedcolumns
    \item[\augmark] A TTS model is used to augment the training data. 
    \item[\firemark] The number of trainable parameters.
    \item[\zeroshotmark] The LLM is used zero-shot for summary generation.

    \end{multicols}
    
\end{tablenotes}
\end{threeparttable}
}
\caption{Quantitative synthesis of cascaded and end-to-end speech summarization models on the \protect\hyperlink{how2-row}{How2} dataset, comparing architectures, input settings, parameter counts, and reported performance (ROUGE-L and BERTScore).}
\label{tab:quantitative}
\end{table*}

%% file: 07_challenges.tex
\section{Critical Gaps and Future Directions}
\label{sec:future}

\paragraph{Limited Reliability of Evaluation.} A key bottleneck remains the lack of trustworthy evaluation practices for SSum. Most existing datasets rely on surrogate summaries, often lack audio data, and are limited by availability\footnote{Most E2E approaches presented in Section \ref{sec:end2end} are exclusively benchmarked on How2, a dataset that is now unavailable and based on surrogate summaries.}. The majority also focus solely on English, restricting broader applicability. Simultaneously, ROUGE remains the dominant metric, despite its limited suitability for SSum. While LLM-based judges are gaining traction, common evaluation protocols are lacking. Human evaluations are often incomparable due to differences in setups, and few approaches account for speech-specific phenomena such as disfluencies, speaker variation, and background noise.

\paragraph{Personalization and Controllability.} Summary needs vary by domain, audience, and intent. As \citet{tuggener_are_2021} outline, meeting summaries alone span formats from action items to narrative recaps, highlighting the mismatch between surrogate summaries and real user needs. Future work should enable controllable summarization along dimensions like length, focus, or style, and support personalization to user roles or preferences.

\paragraph{Multilingual and Cross-Lingual SSum.} 
Research on cross-lingual SSum is still in its early stages. On the dataset side, first works have begun to construct cross-lingual resources by translating references \citep{koneru-etal-2025-kits,papi2025mcifmultimodalcrosslingualinstructionfollowing}, and the task has also been featured in recent evaluation campaigns \citep{agostinelli-etal-2025-findings}. Other work has leveraged cross-lingual TSum datasets by injecting typical ASR errors to simulate transcripts, which are then summarized \citep{pontes-et-al-2019-cross-lingual}.
Modeling efforts have mostly focused on cascaded setups with an intermediate MT module \cite{nelson_cross-lingual_2024} or on integrated models that jointly translate and summarize \cite{kano2023summarize}, yet E2E settings remain largely untapped. 

Closely related, multilingual SSum has likewise received limited attention. Most datasets rely on English speech (\cref{tab:datasets}), with only a few resources covering non-English (\cref{tab:nonenglish-datasets}). Some corpora do provide naturally occurring speech–summary pairs in multiple languages, such as the \hyperlink{spotify-row}{Spotify Podcast Dataset} and the \hyperlink{ELITR-row}{ELITR Minuting Corpus}, but such resources remain the exception. More recently, \citet{chen_floras} constructed summaries across 50 languages by combining LLM-based pseudo-labeling with selective human verification.

\paragraph{Underexplored Frontiers.} Several promising directions in SSum remain underexplored. Online and real-time summarization has seen limited work, with only a few streaming-capable approaches \cite{le-duc_real-time_2024,schneider2025policiesevaluationonlinemeeting}. Multi-document or multi-source SSum, where models process multiple speech inputs or supplemental materials, is also rare despite its relevance in collaborative settings \cite{kirstein_tell_2024}.


%% file: 08_conclusion_limitation.tex
\section{Conclusion}
Despite the progress made in speech summarization, challenges remain, particularly in developing multilingual datasets and evaluation benchmarks that accurately reflect real-world use cases. Future work will need to address these gaps while continuing to refine models for better faithfulness and efficiency. This survey takes a step toward addressing these challenges by providing a comprehensive overview of existing datasets, summarization approaches, and evaluation methods, and by promoting a more holistic view of SSum as a distinct and multifaceted research domain. As the field advances, SSum is poised to play a crucial role in enabling scalable, accessible insights from large, diverse collections of audiovisual content. 

\section*{Limitations}
While we have made efforts to provide a thorough review of the literature on speech summarization, some relevant works may have been overlooked due to variations in search criteria or keywords. Additionally, given the scope of this survey, we focus on the high-level aspects of the approaches and do not delve into an exhaustive, detailed experimental comparison. It is also worth noting that the field is evolving rapidly with the recent emergence of all-purpose language models. While we present these advancements, the widespread adoption of such models may significantly alter the landscape of speech summarization in the near future.

\section*{Ethical Considerations}

Although several critical issues related to AI systems, such as bias, explainability, and fairness, have received increasing attention in recent work \citep{mei-etal-2023-foveate, brandl-etal-2024-interplay, gallegos-etal-2024-bias}, SSum remains a comparatively underexplored area \citep{liu-etal-2023-responsible}. Recent research has begun to highlight the gap in assessing its ethical, legal, and societal implications \citep{shandilya2021fair, keswani2021dialects, merine2022risks, steen-markert-2024-bias}.

Further, fairness concerns emerge when summaries do not equally represent content across demographic groups \citep{Dash2019fairness}. These challenges are exacerbated by the upstream limitations of ASR: performance gaps across accents and socio-economic status \citep{asrforreal2021}, the impact of disfluencies on syntactic and semantic accuracy \citep{mujtaba-etal-2024-lost,teleki_etal_2024_quantifying}, and subtle stereotypical tendencies in spoken LLMs \citep{li2024spoken}. Such errors not only degrade transcription quality but also propagate into the summary, compounding downstream biases \citep{sharma-etal-2024-speech}.

Lastly, SSum systems are active media agents that selectively extract and re-present information from audio or video sources, condensing spoken content into a more concise or structured written summary. In doing so, SSum serves as a powerful tool for controlling the selection and presentation of knowledge. These dynamics raise important questions about the broader consequences of algorithmic and engineering decisions, especially regarding how meaning is conveyed, distorted, or lost. The societal impact of automated summaries goes beyond sensitive domains like medicine, where inaccuracies could lead to misdiagnosis or harmful health outcomes \citep{otmakhova-etal-2022-patient}. Also in fields like scientific communication or news reporting, fluent but incorrect summaries can mislead and misinform \citep{zhao-etal-2020-reducing}. These risks are further amplified in speech summarization, where disfluencies, ambiguity, and the lack of structural cues in spoken language make faithful abstraction especially challenging \citep{kirstein_cads_2025}. As language models become increasingly fluent and persuasive, the threat of confidently wrong summaries becomes all the more pressing.





%% file: appendices/history_appendix.tex
\shortlong{
\section{Early Work}\label{app:history}
\input{sections/history}
}
{}

\shortlong{
\section{Adjacent Speech-to-Text Tasks}\label{app:adjacent}
\input{sections/adjacent_stt}
}
{}

%% file: appendices/datasets_appendix.tex
\onecolumn

\section{Datasets}

\subsection{Non-English Datasets}


\begin{table*}[!h]
\centering
\renewcommand{\arraystretch}{0.8}
\setlength{\aboverulesep}{1.5pt}
\setlength{\belowrulesep}{1.5pt}
\setlength{\tabcolsep}{0.4em}
\renewcommand{\baselinestretch}{0.95}
\newcommand{\customtoprule}{\specialrule{1pt}{0pt}{2pt}}
\newcommand{\custombottomrule}{\specialrule{1pt}{2pt}{0pt}}
\newcommand{\custommidrule}{\specialrule{0.3pt}{1.5pt}{1.5pt}}
\newcommand{\seprule}{\specialrule{1pt}{2pt}{2pt}}
\tiny
\begin{threeparttable}
\begin{tabular}{P{0pt} P{1.35cm}  P{1.70cm} P{1.85cm} P{0.5cm} P{1.60cm} P{2.40cm} C{0.875cm} C{0.5cm} C{0.5cm} C{1.05cm}}
\customtoprule
& \textbf{Dataset} & \textbf{Reference} & \textbf{Domain} & \textbf{Lang.} & \textbf{Size} & \textbf{Summary Type} & \textbf{Transcript} & \textbf{Audio} & \textbf{Video} & \textbf{License}\\
\custommidrule
& \href{https://clrd.ninjal.ac.jp/csj/en/}{CSJ~\faExternalLink*} 
& \citet{maekawa_corpus_2003} 
& Academic speech (various types) 
& JA 
& 3.3k recordings (661 hours) 
& Abstractive \& Extractive 
& Manual 
& \cmark 
& \xmark 
& Paid \\
\custommidrule
& \href{https://github.com/hahahawu/VCSum}{VCSum~\faExternalLink*} 
& \citet{wu_vcsum_2023} 
& Roundtable meetings (from Chinese video-sharing websites) 
& ZH 
& 239 meetings (230 hours) 
& Abstractive (overall, segment-level, and chapter titles) \& Extractive 
& ASR 
& \xmark 
& \xmark 
& MIT \\
\custommidrule
& \href{https://github.com/farahadeeba/CLEMeetingCorpus}{CLE Meeting Corpus~\faExternalLink*} 
& \citet{sadia_meeting_2024} 
& Administrative \& technical meetings (virtual, mostly scenario-driven) 
& UR 
& 240 meetings 
& Abstractive (overall summaries, multiple) 
& Manual 
& \xmark 
& \xmark 
& \qmark\tnote{h} \\
\custommidrule
& \href{https://huggingface.co/datasets/MERaLiON/Multitask-National-Speech-Corpus-v1}{MNSC~\faExternalLink*}
& \citet{he_meralion-audiollm_2025} 
& Conversations of various nature (IMDA NSC Corpus)
& SGE 
& \textasciitilde 100 hours
& Abstractive 
& Manual 
& \cmark 
& \xmark 
& Singapore Open Data License \\
\seprule
& \href{https://huggingface.co/datasets/leduckhai/VietMed-Sum}{VietMed-Sum~\faExternalLink*} 
& \citet{le-duc_real-time_2024} 
& Medical conversations 
& VI 
& 16 hours 
& Abstractive (local \& global) 
& Manual 
& \cmark 
& \xmark 
& \qmark\tnote{a} \\
\custombottomrule
\end{tabular}
\begin{tablenotes}[flushleft]\tiny\parindent=1em
        \item[\text{\makebox[0.5em][l]{a}}] \makebox[0.8em][c]{\qmark} ~\strut No explicit license has been provided.
\end{tablenotes}
\end{threeparttable}
\caption{Non-English Datasets Related to the Speech Summarization Task}
\label{tab:nonenglish-datasets}
\end{table*}

\subsection{Chat-Based Datasets}

\begin{table*}[!h]
\centering
\renewcommand{\arraystretch}{0.8}
\setlength{\aboverulesep}{1.5pt}
\setlength{\belowrulesep}{1.5pt}
\setlength{\tabcolsep}{0.4em}
\renewcommand{\baselinestretch}{0.95}
\tiny
\newcommand{\customtoprule}{\specialrule{1pt}{0pt}{2pt}}
\newcommand{\custombottomrule}{\specialrule{1pt}{2pt}{0pt}}
\newcommand{\custommidrule}{\specialrule{0.3pt}{1.5pt}{1.5pt}}
\newcommand{\seprule}{\specialrule{1pt}{2pt}{2pt}}
\begin{threeparttable}
\begin{tabular}{P{0pt} P{1.6cm} P{1.7cm} P{2.2cm} P{0.5cm} P{1.25cm} P{2.1cm} C{0.875cm} C{0.5cm} C{0.5cm} C{1.1cm}}
\customtoprule
& \textbf{Dataset} & \textbf{Reference} & \textbf{Domain} & \textbf{Lang.} & \textbf{Size} & \textbf{Summary Type} & \textbf{Transcript} & \textbf{Audio} & \textbf{Video} & \textbf{License} \\
\custommidrule
& \href{https://github.com/guyfe/Tweetsumm}{TweetSumm~\faExternalLink*} 
& \citet{feigenblat_tweetsumm_2021} 
& Customer service chats (Twitter) 
& EN 
& 1.1k dialogues 
& Abstractive \& Extractive (multiple) 
& N/A (Chat) 
& \xmark 
& \xmark 
& CDLA-Sharing-1.0 \\
\custommidrule
& \href{https://github.com/xiaolinAndy/CSDS}{CSDS~\faExternalLink*} 
& \citet{lin_csds_2021} 
& Customer service chats (JD.com) 
& ZH 
& 2.5k dialogues 
& Extractive \& Abstractive (role-oriented, topic-structured, multiple) 
& N/A (Chat) 
& \xmark 
& \xmark 
& \qmark\tnote{a} \\
\seprule
& \href{https://huggingface.co/datasets/Samsung/samsum}{SAMSum~\faExternalLink*} 
& \citet{gliwa_samsum_2019} 
& Chat conversations (scenario-driven) 
& EN 
& 16k dialogues 
& Abstractive 
& N/A (Chat) 
& \xmark 
& \xmark 
& CC-BY-NC-ND-4.0 \\
\seprule
& \href{https://github.com/synlp/HET-MC/tree/main/data_preprocessing}{MC~\faExternalLink*} 
& \citet{song_etal_2020_summarizing} 
& Medical conversations (Chunyu Yisheng)
& ZH 
& 16 hours 
& Abstractive (local \& global) 
& N/A (Chat) 
& \xmark   
& \xmark 
& \qmark\tnote{a} \\
\custombottomrule
\end{tabular}
\begin{tablenotes}[flushleft]\tiny\parindent=1em
        \item[\text{\makebox[0.5em][l]{a}}] \makebox[0.8em][c]{\qmark} ~\strut No explicit license has been provided.
\end{tablenotes}
\end{threeparttable}
\caption{Chat-Based Summarization Datasets Structurally Similar to Speech}
\label{tab:chat-datasets}
\end{table*}

\subsection{Dataset Derivatives and Augmentations}

\begin{table*}[!h]
\centering
\renewcommand{\arraystretch}{0.8}
\setlength{\aboverulesep}{1.5pt}
\setlength{\belowrulesep}{1.5pt}
\setlength{\tabcolsep}{0.4em}
\renewcommand{\baselinestretch}{0.95}
\tiny
\newcommand{\customtoprule}{\specialrule{1pt}{0pt}{2pt}}
\newcommand{\custombottomrule}{\specialrule{1pt}{2pt}{0pt}}
\newcommand{\custommidrule}{\specialrule{0.3pt}{1.5pt}{1.5pt}}
\newcommand{\seprule}{\specialrule{1pt}{2pt}{2pt}}
\begin{threeparttable}
\begin{tabular}{P{0pt} P{1.6cm} P{1.7cm} P{1.85cm} P{1.5cm} P{4.8cm} C{1.05cm}}
\customtoprule
& \textbf{Dataset} & \textbf{Reference} & \textbf{Base Dataset} & \textbf{Lang.} & \textbf{Extension Type} & \textbf{License} \\
\custommidrule
& \href{https://github.com/Jungjee/AugSumm}{AugSumm~\faExternalLink*} 
& \citet{jung_augsumm_2024} 
& \hyperlink{how2-row}{How2} 
& EN 
& Synthetic summaries generated by GPT-3.5 Turbo (direct + paraphrased) to enrich summary diversity
& \qmark\tnote{a} \\
\custommidrule
& \href{https://github.com/talkiq/dialpad-ai-research/tree/main/tiny_titans}{QMSum-I~\faExternalLink*} 
& \citet{fu_tiny_2024} 
& \hyperlink{qmsum-row}{QMSum}
& EN 
& Instruction-based summaries (long, medium, short) generated by GPT-4
& \qmark\tnote{a} \\
\custommidrule
& \href{https://github.com/hkim-etri/ExplainMeetSum}{ExplainMeetSum~\faExternalLink*} 
& \citet{kim_explainmeetsum_2023} 
& \hyperlink{qmsum-row}{QMSum}
& EN 
& Annotated evidence sentences in the transcript that faithfully support sentences in the summary
& MIT \\
\custommidrule
& \href{https://github.com/psunlpgroup/MACSum}{MACSum~\faExternalLink*}
& \citet{zhang-etal-2023-macsum}
& \hyperlink{qmsum-row}{QMSum} \& CNN/DM
& EN
& Human-annotated summaries with mixed attributes (length, extractiveness, specificity, topic, speaker); includes evidence spans and summary titles
& CC-BY-NC-SA 4.0 \\
\custommidrule
& \href{https://github.com/FKIRSTE/emnlp2024-personalized-meeting-sum}{MS-AMI~\faExternalLink*} 
& \citet{kirstein_tell_2024} 
& \hyperlink{ami-row}{AMI}
& EN 
& Enriches the source data with processed, supplementary materials (whiteboard drawings, slides, notes) using GPT-4o and Aspose for text extraction
& Apache-2.0 \\
\custommidrule& \href{https://huggingface.co/datasets/retkowski/length-controllability-evaluation}{YTSeg-LC~\faExternalLink*} 
& \citet{retkowski_zero-shot_2025} 
& \hyperlink{ytseg-row}{YTSeg} 
& EN 
& Length-controlled summaries generated by LLaMA 3 and other LLMs
& CC-BY-NC-SA 4.0 \\
\custommidrule
\incmark\tnote{a}
& \href{https://huggingface.co/datasets/microsoft/MeetingBank-QA-Summary}{MeetingBank-QA-Summary~\faExternalLink*} 
& \citet{pan_llmlingua-2_2024} 
& \hyperlink{meetingbank-row}{MeetingBank} 
& EN 
& The test set is enriched by summaries and question-answer pairs for each transcript generated by GPT-4
& CC-BY-NC-SA 4.0 \\
\custommidrule
\incmark\tnote{a}
& \href{https://huggingface.co/datasets/microsoft/MeetingBank-LLMCompressed}{MeetingBank-LLMCompressed~\faExternalLink*} 
& \citet{pan_llmlingua-2_2024} 
& \hyperlink{meetingbank-row}{MeetingBank} 
& EN 
& Enriches the train data split by chunk-level compressed meeting transcripts generated by GPT-4 
& CC-BY-NC-SA 4.0 \\
\custommidrule
\incmark\tnote{a}
& \href{https://github.com/amazon-science/tofueval}{TofuEval~\faExternalLink*} 
& \citet{tang_tofueval_2024} 
& \hyperlink{meetingbank-row}{MeetingBank} \& \hyperlink{mediasum-row}{MediaSum}
& EN 
& Expert annotations of topic-focused LLM summaries on factual consistency and completeness
& MIT-0 \\
\custombottomrule
\end{tabular}
\begin{tablenotes}[flushleft]\tiny\parindent=1em
        \item[\text{\makebox[0.5em][l]{a}}] \makebox[0.8em][c]{\qmark} ~\strut No explicit license has been provided.
        \item[\text{\makebox[0.5em][l]{b}}] \makebox[0.8em][c]{\incmark} ~\strut Not all data partitions were augmented.
\end{tablenotes}
\end{threeparttable}
\caption{Derivatives of and Augmentations to Existing Speech Summarization Sources}
\label{tab:dataset-derivatives}
\end{table*}

%% file: appendices/eval_appendix.tex
\clearpage
\shortlong{
\twocolumn
\section{Evaluation of Speech Summaries}\label{app:eval}
\subsection{Metrics Borrowed From TSum}

\paragraph{Lexical Overlap Metrics.}
Lexical overlap metrics assess similarity based on shared surface-level units. ROUGE \citep{lin-2004-rouge}, designed to maximize recall, is the most widely used metric \citep{fabbri_summeval_2021, Sharma2024}, though implementation errors have led to incorrect evaluations in the past \citep{grusky-2023-rogue}. 
Moreover, early work has shown that the presence of disfluencies, multiple speakers, and the lack of structure in spontaneous speech diminish the correlation between ROUGE scores and human judgment \citep{liu-liu-2008-correlation}.
BLEU \citep{papineni-etal-2002-bleu, post-2018-call} and METEOR \citep{banerjee-lavie-2005-meteor}, remain common to evaluate summaries despite being developed for machine translation.
Methods like Basic Elements \citep{hovy-etal-2006-automated} and the Pyramid Method \citep{nenkova-passonneau-2004-evaluating} improve overlap metrics by also considering syntactic dependencies and content units.

Despite their efficiency, these lexical overlap metrics struggle to evaluate the faithfulness to the input \citep{bhandari-etal-2020-metrics, maynez-etal-2020-faithfulness, wang-etal-2020-asking}, fail to distinguish similar or high-scoring candidates \citep{peyrard-2019-studying, bhandari-etal-2020-metrics}, and are often outperformed by model-based evaluators, which has been shown for dialog summarization by \citet{gao-wan-2022-dialsummeval}. Moreover, since they do not use the source, i.e., speech or transcript, they often fail to account for SSum-specific attributes.

\paragraph{Embedding-Based Metrics.}
Embedding-based metrics capture semantic similarity through sentence or token embeddings. Yet, they still struggle to assess factual accuracy, fully capture shared information \citep{deutsch-roth-2021-understanding}, and distinguish similar candidates \citep{bhandari-etal-2020-metrics}. 

BERTScore \citep{zhang_bertscore_2020}, one of the most prominent embedding-based metrics, compares contextualized token embeddings between the summary and reference. Yet, \citet{kirstein_whats_2024} find that BERTScore has not been tested for meeting summarization and is often unsuitable due to its 512-token context limit, which is frequently exceeded by lengthy transcripts. Other model-based evaluators include MoverScore \citep{zhao-etal-2019-moverscore}, which measures the earth-mover distance between embeddings, capturing both content overlap and divergence and SPEEDScore \citep{akula_sentence_2022}, which evaluates summary efficiency by balancing compression and information retention using sentence-level embeddings.

\paragraph{Trained Evaluators.}
Recent approaches have focused on training models for more holistic summary evaluation \citep{NEURIPS2021_e4d2b6e6, zhong_towards_2022}, as well as for specific dimensions like factual accuracy \citep{kryscinski_evaluating_2020, wang-etal-2020-asking, durmus-etal-2020-feqa, scialom-etal-2021-questeval}.  Other models refine evaluations using counterfactual estimation \citep{xie-etal-2021-factual-consistency} and causal graphs \citep{ling-etal-2025-enhancing}. However, even evaluation-specific models, particularly reference-free ones, may be prone to spurious correlations such as summary length \citep{durmus_spurious_2022}.

\onecolumn}{\section{Evaluation of Speech Summaries}\label{app:eval}}

\subsection{Usage of Speech Summarization Metrics over Time}
\begin{figure*}[!ht]
    \centering
    \includegraphics[trim=0mm 0mm 0mm 13mm, clip, width=0.8\textwidth]{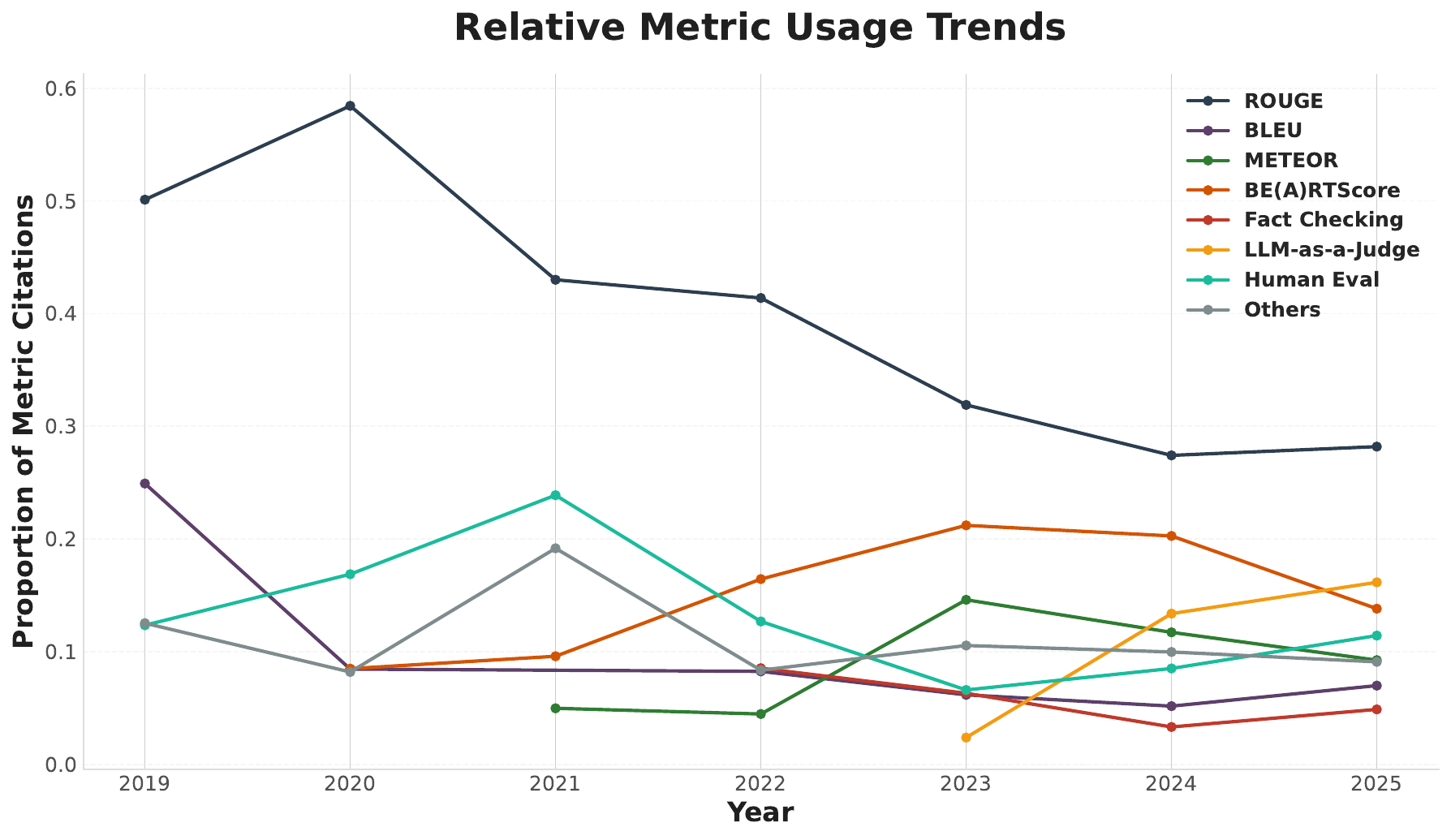}
    \caption{Proportion of citations for different evaluation metrics over time (based on the SSum papers included in this survey; after 2018), normalized by the total number of citations per year. \textit{Others} includes all metrics with three or fewer citations.}
    \label{fig:metrics_over_years}
\end{figure*}
\noindent
    \begin{minipage}[t]{0.48\textwidth}
    \paragraph{Trends in Usage of Metrics.} \cref{fig:metrics_over_years} shows the normalized proportion of citations for various evaluation metrics from 2019 to 2025. We observe an increase in the use of metrics other than ROUGE \citep{lin-2004-rouge} from 2020 to 2024, followed by stabilization in 2025. BE(A)RTScore (BERTScore, [\citealp{zhang_bertscore_2020}] and BARTScore [\citealp{NEURIPS2021_e4d2b6e6}]) grows steadily from 2020 to 2023 but starts to lose popularity since then. Human evaluation has remained relatively stable throughout the years. By 2025, LLM-as-a-Judge becomes the second most used metric, emerging in 2023 and rapidly gaining popularity. A detailed overview of the different LLM-as-a-Judge methods can be found in \cref{tab:llm_judge}, and a detailed overview of different human evaluation approaches can be found in \cref{tab:human_eval}.
    \phantom{Line 1}\\
    \paragraph{Fact Checking.} The \textit{Fact Checking} category includes the following metrics: FactCC \citep{huang_swing_2023}, QUALS \citep{huang_swing_2023}, QAGS \citep{wang-etal-2020-asking, suresh_diasynth_2025, manakul_podcast_2022}, QAFactEval \citep{huang_swing_2023, tang_tofueval_2024}, FactVC \citep{liu-wan-2023-models}, SummaCConv \citep{laban-etal-2022-summac}, FACTSCORE \citep{min_etal_2023_factscore} and QAEval \citep{deutsch-etal-2021-towards, hu_meetingbank_2023}.
    \end{minipage}\hfill
    \begin{minipage}[t]{0.48\textwidth}

        \paragraph{Others.} The \textit{Others} category includes metrics less frequently used for speech summarization, such as F-score \citep{lv_vt-ssum_2021, palaskar_multimodal_2019}, Perplexity \citep{kirstein-etal-2024-whats, retkowski_zero-shot_2025},  ChrF \citep{popovic-2015-chrf,jorgensen_cross-lingual_2025}, Sentence Cosine Similarity \citep{li_hierarchical_2021}, BoC (Bag of Characters; \citealt{chen_summscreen_2022}), BLANC \citep{kirstein-etal-2024-whats, vasilyev-etal-2020-fill}, LENS \citep{maddela-etal-2023-lens, kirstein-etal-2024-whats}, MoverScore \citep{zhao-etal-2019-moverscore, hu_meetingbank_2023}, CIDEr \citep{vedantam2015cider, qiu_mmsum_2024}, and SPICE \citep{spice2016, qiu_mmsum_2024}.
    \end{minipage}

\newpage
\onecolumn
\subsection{LLM-as-a-Judge for Speech Summarization}
\input{tables/llm_judge_table}

\newpage
\onecolumn
\subsection{Human Evaluation for Speech Summarization}
\input{tables/human_eval_table}

%% file: tables/llm_judge_table.tex
\newcommand{\customtoprule}{\specialrule{1pt}{0pt}{2pt}}
\newcommand{\custombottomrule}{\specialrule{1pt}{2pt}{0pt}}
\newcommand{\seprule}{\specialrule{1pt}{2pt}{2pt}}

\begin{table}[!h]
    \centering
    \renewcommand{\arraystretch}{0.75}
    \renewcommand{\baselinestretch}{0.95}
    \setlength{\aboverulesep}{1.8pt}
    \setlength{\belowrulesep}{1.8pt}
    \resizebox{\linewidth}{!}{%
    \begin{tabular}{ P{1.85cm} P{4.5cm} P{5.5cm} P{2.5 cm} P{3.0 cm} P{1cm}}
    \customtoprule
    Method & Judge Model & Criteria (Framework) & Data & Reference & Total \\
    \seprule
    \multirow[t]{7}{*}{\parbox{1.35cm}{\vspace{27pt}Absolute Score/Scale}}
    & Llama-3.1-8B-Instruct \citep{grattafiori2024llama3herdmodels} & Relevance, Coherence, Conciseness, Factual Accuracy & Output Summary, Reference & \citet{zufle_nutshell_2025} &  7\\
    \cdashlinelrg{2-5}
    &Llama-3.1-8B-Instruct \citep{grattafiori2024llama3herdmodels} & General Alignment with Reference & Output Summary, Reference & \citet{zufle_nutshell_2025} \\
    \cdashlinelrg{2-5}
     &  Meta-Llama-3-70B  \citep{grattafiori2024llama3herdmodels}  & Content, Accuracy, and Relevance & Output Summary, Reference &  \citet{he_meralion-audiollm_2025} \\
    \cdashlinelrg{2-5}
    & GPT-4 \citep{openai2024gpt4technicalreport} & Overall Quality, Instruction Adherence & Transcript, Output Summary  &  \citet{microsoft_phi-4-mini_2025} \\
\cdashlinelrg{2-5}

    & Prometheus-8x7B \citep{Brazil2019PROMETHEUS} & Honesty, Factual Validity, Completeness \citep[Prometheus-Eval,][]{Brazil2019PROMETHEUS} & Transcript, Output Summary & \citet{thulke_prompting_2024} \\ 

    \cdashlinelrg{2-5}
     & GPT-4o \citep{openai2024gpt4technicalreport} & Redundancy, Incoherence, Language, Omission, Coreference, Hallucination, Structure, Irrelevance  \citep[MESA,][]{kirstein_is_2025}   & Transcript, Output Summary & \citet{kirstein_is_2025} \\

     \cdashlinelrg{2-5}
    & GPT-4-32k  \citep{openai2024gpt4technicalreport} &  Adequacy, Relevance, Topicality, Fluency, Grammaticality & Transcript, Output Summary & \citet{ghosal_overview_2023} \\
    \seprule
    \multirow[t]{2}{*}{\parbox{1.35cm}{\vspace{0pt}Ranking }}
    & Llama-3.1-8B-Instruct  \citep{grattafiori2024llama3herdmodels}  & Relevance, Coherence, Conciseness, Factual Accuracy & Output Summaries, Reference & \citet{zufle_nutshell_2025} & 2\\
    \cdashlinelrg{2-5}
    & GPT-based models \citep{openai2024gpt4technicalreport} & Completeness, Conciseness, Factuality  \citep[CREAM,][]{gong_cream_2024} & Output Summaries & \citet{gong_cream_2024} \\
    \seprule
    \multirow[t]{2}{*}{\parbox{1.35cm}{\vspace{27pt}Pairwise Comparison}}
    & GPT4-Turbo \citep{openai2024gpt4technicalreport} & Not Specified & Output Summary, Reference & \citet{matsuura_sentence-wise_2024} & 2\\
    \cdashlinelrg{2-5}
    & GPT-4o \citep{openai2024gpt4technicalreport}& General Performance \citep[Alpaca Eval,][]{dubois_alpacafarm_2023} & Transcript, Output Summary, Baseline Summary & \citet{retkowski_zero-shot_2025} \\
    \seprule

    \multirow[t]{3}{*}{\parbox{1.35cm}{\vspace{0pt}Accuracy}}
    & GPT-4 \citep{openai2024gpt4technicalreport} & Hallucination & Transcript, Output Summary  &  \citet{microsoft_phi-4-mini_2025} & 3 \\
    \cdashlinelrg{2-5}
    & GPT-4o \citep{openai2024gpt4technicalreport} & Faithfulness, Completeness, Conciseness \citep[FineSureE,][]{song-etal-2024-finesure} & Transcript, Output Summary & \citet{thulke_prompting_2024} \\ 

    \cdashlinelrg{2-5}
     &GPT-4 \citep{openai2024gpt4technicalreport} among other, weaker judges & Factual Correctness  & Transcript, Output Summary/Sentence & \citet{tang_tofueval_2024} \\

    \custombottomrule
   
    \end{tabular}%
    }
    \caption{Different ways of LLM-as-a-Judge for SSum, based on the SSum papers included in this survey.}
    \label{tab:llm_judge}
\end{table}

%% file: tables/human_eval_table.tex
\begin{table}[!h]
    \centering
    \renewcommand{\arraystretch}{0.75}
    \renewcommand{\baselinestretch}{0.95}
    \setlength{\aboverulesep}{1.8pt}
    \setlength{\belowrulesep}{1.8pt}
    \resizebox{\linewidth}{!}{%
    \begin{tabular}{ P{1.35cm} P{3cm} P{6.0cm} P{4.5 cm} P{3.5 cm} P{1cm}}
    \customtoprule
    Method & Annotators & Criteria & Data & Reference & Total \\
    \seprule
    \multirow[t]{12}{*}{\parbox{1.35cm}{\vspace{13pt}Likert Scale}}
    & Crowdsourced & Readability, Relevance & Transcript, Output & \citet{zhu_hierarchical_2020} & 13\\
    \cdashlinelrg{2-5}
    & Crowdsourced & Informativeness, Relevance, Coherence & Video, Output, Reference & \citet{palaskar_multimodal_2019} & \\
    \cdashlinelrg{2-5}
    & Crowdsourced & Informativeness, Redundancy & Transcript, Output & \citet{song-etal-2022-towards}  & \\
    \cdashlinelrg{2-5}
    & Crowdsourced & Informativeness, Factuality, Fluency, Coherence, Redundancy & Video, Transcript, Output & \citet{hu_meetingbank_2023}  & \\
    \cdashlinelrg{2-5}
    & Graduate Students & Frequency of Transcript Challenges, Error Quality Impact & Transcripts, Output, Reference & \citet{kirstein-etal-2024-whats}  & \\
    \cdashlinelrg{2-5}
    & Domain Experts & Adequacy, Fluency, Relevance & Transcript, Output & \citet{schneider2025policiesevaluationonlinemeeting}  & \\
    \cdashlinelrg{2-5}
    & Not Specified & Fluency, Coherence, Factual Consistency & Not Specified & \citet{fu_tiny_2024} \\
    \cdashlinelrg{2-5}
    & Annotators with English Expertise & Readability, Conciseness, Coverage & Transcript, Output, Reference & \citet{zhang_summn_2022}  & \\ 
    \cdashlinelrg{2-5}
    & Domain Experts & Fluency, Consistency, Relevance, Coherence & Transcript, Output & \citet{le-duc_real-time_2024}  & \\
    \cdashlinelrg{2-5}
    & Graduate Students & Error Types Detection & Transcript, Output & \citet{kirstein_is_2025} \\
    \cdashlinelrg{2-5}
    & Not Specified & Fluency, Consistency, Relevance, Coherence & Source (Dialog), Output & \citet{chen_dialogsum_2021}  & \\
    \cdashlinelrg{2-5}
    & Experienced Annotators & Adequacy (Informativeness), Fluency, Grammatical Correctness, Relevance & Transcripts, Output & \citet{ghosal_overview_2023}  & \\
    \cdashlinelrg{2-5}
    & Well-Educated Volunteers & Informativeness, Redundancy, Fluency, Matching Rate & Transcripts, Output & \citet{lin_csds_2021}  & \\
    \seprule
    \multirow[t]{2}{*}{\parbox{1.35cm}{\vspace{27pt}Best-Worst Scaling}}
    & Domain Experts & Relevance, Coherence, Conciseness, Factual Accuracy & Outputs, Reference & \citet{zufle_nutshell_2025} & 2\\
    \cdashlinelrg{2-5}
    & Graduate Students & Fluency, Informativeness, Faithfulness & Source (Dialog), Outputs & \citet{zhong_dialoglm_2022}  & \\
    \seprule
    \multirow[t]{5}{*}{\parbox{1.35cm}{\vspace{27pt}Pairwise Comparison}}
    & Crowdsourced & Coherence, Informativeness, Overall quality & Transcript, Outputs & \citet{cho_streamhover_2021}  & 5 \\ 
    \cdashlinelrg{2-5}
    & Crowdsourced & Factual Consistency, Informativeness & Source (Dialog), Outputs & \citet{zhu_factual_2025}  & \\ 
    \cdashlinelrg{2-5}
    & Crowdsourced & Recall, Precision, Faithfulness & Source (Dialog), Outputs & \citet{huang_swing_2023} \\
    \cdashlinelrg{2-5} 
    & Not Specified & Not Specified & Not Specified & \citet{eom_squba_2025}  & \\ 
    \cdashlinelrg{2-5}
    & Crowdsourced Domain Experts & Readability, Informativeness & Outputs & \citet{feigenblat_tweetsumm_2021}  & \\ 
    \seprule
    \multirow[t]{6}{*}{\parbox{1.35cm}{\vspace{27pt}QA-Based\\Eval}}
    & Crowdsourced & Podcast Specifics, Language, Redundancy & Transcript, Output & \citet{song-etal-2022-towards} & 6\\
    \cdashlinelrg{2-5}
    & Graduate Students & Challenges in Transcript & Transcripts, Output Reference & \citet{kirstein-etal-2024-whats}  & \\
    \cdashlinelrg{2-5}
    & System Users & Comprehension & Audio, Output & \citet{koumpis_automatic_2005}  & \\
    \cdashlinelrg{2-5}
    & Not Specified & Informativeness, Factual Accuracy & Transcripts or Output & \citet{zechner_diasumm_2000}  & \\
    \cdashlinelrg{2-5}
    & Graduate Students & Grammatical Correctness, Semantic Comprehensibility & Audio, Transcript, Output & \citet{li_hierarchical_2021}  & \\
    \cdashlinelrg{2-5}
    & Crowdsourced Domain Experts & Informativeness, Saliency, Readability & Transcripts, Output & \citet{feigenblat_tweetsumm_2021}  & \\
    \seprule
    \multirow[t]{2}{*}{\parbox{1.35cm}{\vspace{16pt}MOS Score}}
    & System Users & Not Specified & Subset of Transcript, Output & \citet{koumpis_automatic_2005} & 2 \\
    \cdashlinelrg{2-5}
    & Not Specified & Relevance & Transcript, Outputs & \citet{chowdhurytranscripterrors2024} & \\
    \seprule
    \multirow[t]{2}{*}{\parbox{1.35cm}{\vspace{1pt}Accuracy}}
    & Not Specified & Readability & Sentences of Output & \citet{banerjee_generating_2015} & 2 \\
    \cdashlinelrg{2-5}
    & Domain Experts & Factual Accuracy & Transcript, Sentences of Output & \citet{tang_tofueval_2024}  & \\
    \seprule
    \multirow[t]{2}{*}{\parbox{1.35cm}{\vspace{12pt}Absolute Score}}
    & Domain Experts & Relevance, Completeness & Transcript, (Topic,) Output & \citet{tang_tofueval_2024} & 3\\
    \cdashlinelrg{2-5}
    & Not Specified & Discourse Relations, Intent, Coreference & Source (Dialog), Output & \citet{chen_dialogsum_2021} & \\
   \cdashlinelrg{2-5}
    & Undergrad Students in Computer Science & Informativeness, Relevance, Importance, Redundancy, Amount of Summary Space given to Topic, Role of Speaker &  Output, Reference & \citet{liu-liu-2008-correlation} \\
    \custombottomrule
    
    \end{tabular}%
    }
    \caption{Different ways of human evaluation for SSum, based on the SSum papers in this survey.}
    \label{tab:human_eval}
\end{table}

%% file: appendices/models_appendix.tex
\clearpage

\section{Approaches}\label{app:models}
\input{tables/opensource_models_table}

 \shortlong{\subsection{Quantitative Comparisons} \input{tables/quanitative_synthesis}\newpage}{}

%% file: tables/opensource_models_table.tex

\subsection{Open-Source Speech Summarization Models}

\begin{table*}[!h]
\centering
\renewcommand{\arraystretch}{0.85}
\setlength{\aboverulesep}{1.5pt}
\setlength{\belowrulesep}{1.5pt}
\setlength{\tabcolsep}{0.4em}
\renewcommand{\baselinestretch}{0.95}
\tiny
\newcommand{\custommidrule}{\specialrule{0.3pt}{1.5pt}{1.5pt}}
\begin{threeparttable}
\begin{tabular}{P{3.0cm} P{2.4cm} P{3.0cm} P{2.7cm} P{1.15cm}}
\customtoprule
\textbf{Model} & \textbf{Reference} & \textbf{Architecture / Backbone} & \textbf{Language / Region Focus} & \textbf{Input Type} \\
\custommidrule
DialogLED (\href{https://huggingface.co/MingZhong/DialogLED-base-16384}{Base \faExternalLink*}, \href{https://huggingface.co/MingZhong/DialogLED-large-5120}{Large \faExternalLink*}) & \citet{zhong_dialoglm_2022} & ED / LED (Longformer) & English (dialogues) & Transcript \\
\href{https://github.com/microsoft/HMNet/blob/main/ExampleInitModel/HMNet-pretrained/README.md}{HMNet~\faExternalLink*} & \citet{zhu_hierarchical_2020} & Hierarchical ED / Transformer & English (meetings) & Transcript \\
\href{https://github.com/psunlpgroup/Summ-N}{Summ-N~\faExternalLink*} & \citet{zhang_summn_2022} & ED / BART & English (dialogues) & Transcript \\
\seprule
\href{https://huggingface.co/Qwen/Qwen2-Audio-7B}{Qwen-2-Audio~\faExternalLink*} & \citet{chu_qwen2-audio_2024} & LLM + Audio Encoder / Qwen + Whisper & Multilingual (EN, ZH, FR, IT, ES, DE, JA) & Speech \\
\href{https://huggingface.co/microsoft/Phi-4-multimodal-instruct}{Phi-4 Multimodal~\faExternalLink*} & \citet{microsoft_phi-4-mini_2025} & LLM + Audio Encoder / Phi-4 + Whisper & Multilingual (EN, ZH, DE, FR, IT, JA, ES, PT) & Speech \\
\seprule
\href{https://huggingface.co/MERaLiON/MERaLiON-AudioLLM-Whisper-SEA-LION}{MERaLiON-AudioLLM~\faExternalLink*} & \citet{he_meralion-audiollm_2025} & LLM + Audio Encoder / SEA-LION V3 + Whisper & Singapore (EN, SGE) & Speech \\
\href{https://huggingface.co/SeaLLMs/SeaLLMs-Audio-7B}{SeaLLMs-Audio~\faExternalLink*} & \citet{SeaLLMs-Audio} & LLM + Audio Encoder / Qwen2-Audio-7B + Qwen2.5-7B-Instruct & Southeast Asia (EN, ZH, ID, TH, VI) & Speech \\
\custombottomrule
\end{tabular}
\end{threeparttable}
\caption{Open-Source Pretrained Models for Summarization from Speech or Speech Transcript Inputs}
\label{tab:speech-models}
\end{table*}

%% file: appendices/venues_appendix.tex
\section{Supplementary Statistics}\label{app:venue}

\begin{figure}[!h]
    \centering
    \centering 
    \includegraphics[trim=0mm 0mm 0mm 13mm, clip, width=0.8\textwidth]{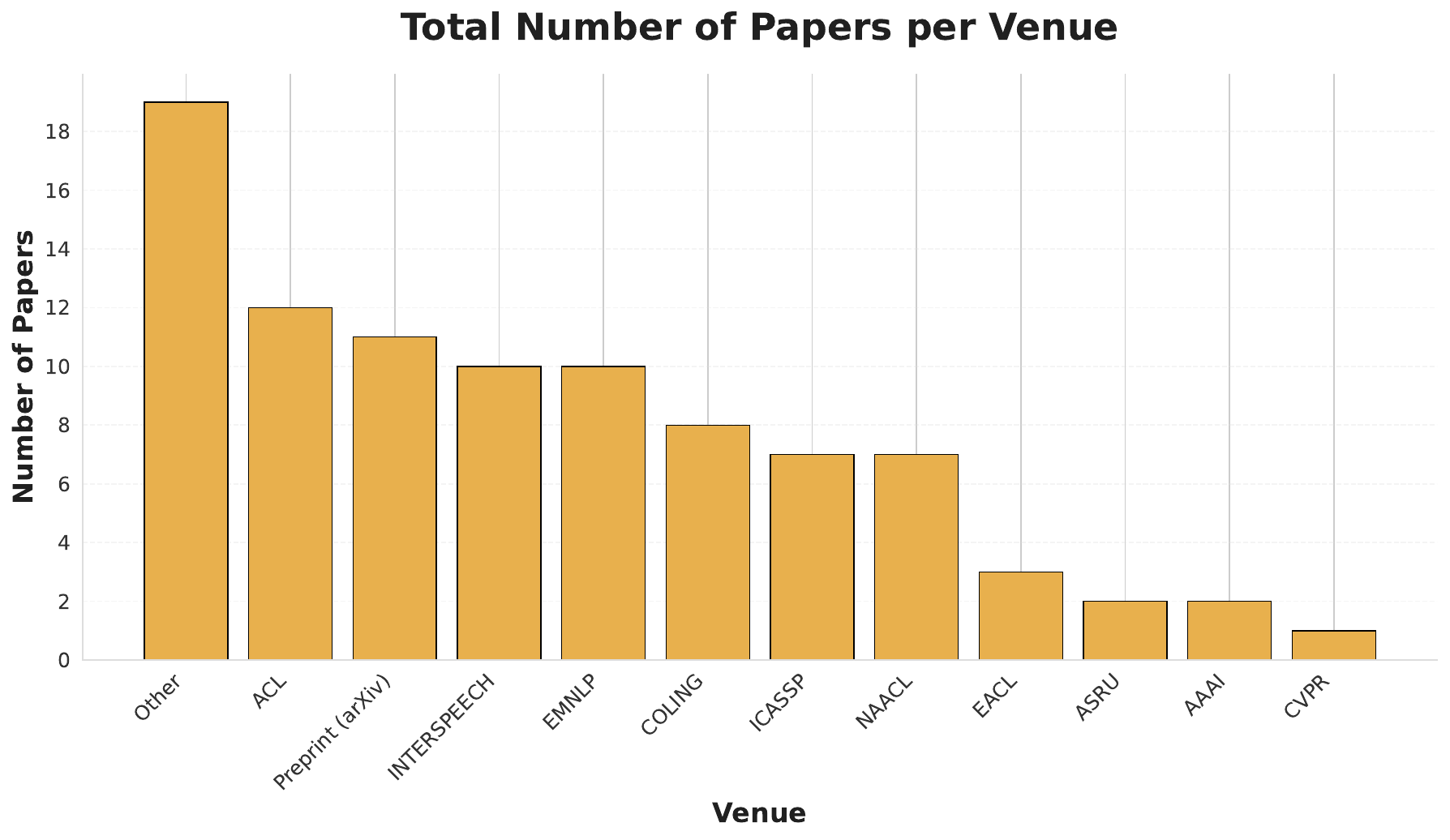}
    \caption{Total number of SSum papers published in different venues, based on the SSum papers included in this survey. Note that papers listed under \textit{Preprint (arXiv)} are only those without a corresponding conference or journal version, avoiding duplication. These papers are largely very recent works or technical reports.}
    \label{fig:venues_over_years}
\end{figure}